\relax
\documentclass[letterpaper]{article} 
\usepackage{times}  
\usepackage{helvet}  
\usepackage{courier}  
\usepackage[hyphens]{url}  
\usepackage{graphicx} 
\usepackage{caption} 
\DeclareCaptionStyle{ruled}{labelfont=normalfont,labelsep=colon,strut=off} 
\usepackage{algorithm}
\usepackage[noend]{algorithmic}
\usepackage{amsmath}
\usepackage{amsfonts}
\usepackage{bbm}
\usepackage{floatflt}
\usepackage{multirow}
\usepackage{enumitem}
\usepackage{tikz}
\usetikzlibrary{positioning}
\usepackage{booktabs}
\usepackage{fullpage}
\usepackage{hyperref}

\newcommand{\qed}{\hfill \ensuremath{\Box}}

\newtheorem{thmx}{Theorem}

\newenvironment{proof}{{\bf Proof:}}{\qed}

\newcommand{\algname}{{\sc LoNe Sampler }}

\tikzset{main node/.style={circle,fill=blue!20,draw,minimum size=1cm,inner sep=0pt},
            }

\usepackage{newfloat}
\usepackage{listings}
\floatstyle{ruled}
\newfloat{listing}{tb}{lst}{}
\floatname{listing}{Listing}
\title{\algname: Graph node embeddings by coordinated local neighborhood sampling}
\author{
Konstantin Kutzkov \\ \normalsize kutzkov@gmail.com
}
\date{}
\begin{document}

\maketitle

\begin{abstract}
Local graph neighborhood sampling is a fundamental computational problem that is at the heart of algorithms for node representation learning. 
Several works have presented algorithms for learning {\em discrete} node embeddings where graph nodes are represented by discrete features such as attributes of neighborhood nodes. Discrete embeddings offer several advantages compared to continuous word2vec-like node embeddings: ease of computation, scalability, and interpretability. We present \algname, a suite of algorithms for generating discrete node embeddings by \underline{Lo}cal \underline{Ne}ighborhood \underline{Sam}pling, and address two shortcomings of previous work. First, our algorithms have rigorously understood theoretical properties. Second, we show how to generate approximate explicit vector maps that avoid the expensive computation of a Gram matrix for the training of a kernel model. Experiments on benchmark datasets confirm the theoretical findings and demonstrate the advantages of the proposed methods. 
\end{abstract}

\section{Introduction}
\label{intro}
Graphs are ubiquitous representation for structured data. They model naturally occurring relations between objects and, in a sense, generalize sequential data to more complex dependencies. Many algorithms originally designed for learning from sequential data are thus generalized to learning from graphs. Learning continuous vector representations of graph nodes, or {\em node embeddings}, have become an integral part of the graph learning toolbox, with applications ranging from link prediction~\cite{node2vec} to graph compression~\cite{compression}. The first algorithm~\cite{deepwalk} for learning node embeddings generates random walks, starting from each node in the graph, and then feeds the sequences of visited nodes into a word embedding learning algorithm such as word2vec~\cite{word2vec}. The approach was extended to a more general setting where random walks can consider different properties of the local neighborhood~\cite{node2vec,line,verse}. An alternative method for training continuous node embeddings is based on matrix factorization of (powers of) the graph adjacency matrix. 

As an alternative, researchers proposed to use {\em coordinated} node sampling for training {\em discrete} node embeddings~\cite{nethash,nodesketch}. In this setting, each sample is an independent estimator of the similarity between nodes. (There are different notions of node similarity but most reflect how easy it is to reach one node from another.)  Thus, sampled nodes themselves can be coordinates of the embedding vectors. We can then compare two node embeddings by their Hamming distance. There are several advantages of discrete embeddings over continuous embeddings. First, we avoid the need to train a (possibly slow) word2vec-like model. Second, the samples are the original graph nodes and contain all meta-information provided in the original input, be it personal data of users of a social network or the weather conditions at railway stations. By sampling, all this information is preserved and this can lead to the design of interpretable algorithms. And finally, the algorithms are truly local and can deal with massive or distributed graphs if only access to the local neighborhood of each node is possible. 

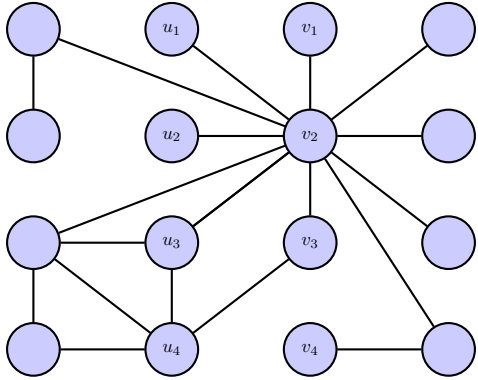
\begin{figure}
\centering
  \begin{tikzpicture}[thick,scale=0.7, every node/.style={transform shape}]
    \node[main node] (y1) {};
    \node[main node] (y2) [below = 1cm  of y1]  {};
    \node[main node] (y3) [below = 1cm  of y2] {};
    \node[main node] (y4) [below = 1cm  of y3] {};
    \node[main node] (u1) [right = 1.6cm  of y1] {$u_1$};
    \node[main node] (u2) [below = 1cm  of u1]  {$u_2$};
    \node[main node] (u3) [below = 1cm  of u2] {$u_3$};
    \node[main node] (u4) [below = 1cm  of u3] {$u_4$};
    
    \node[main node] (v1) [right = 1.6cm  of u1] {$v_1$};
    \node[main node] (v2) [below = 1cm  of v1]  {$v_2$};
    \node[main node] (v3) [below = 1cm  of v2] {$v_3$};
    \node[main node] (v4) [below = 1cm  of v3] {$v_4$};
    
    \node[main node] (z1) [right = 1.6cm  of v1] {};
    \node[main node] (z2) [below = 1cm  of z1] {};
    \node[main node] (z3) [below = 1cm  of z2] {};
    \node[main node] (z4) [below = 1cm  of z3] {};
    

    \path[draw,thick]
    (u2) edge node {} (v2)
    (y1) edge node {} (y2)
    (u1) edge node {} (v2)
    (y3) edge node {} (u3)
    (v3) edge node {} (u4)
    (y4) edge node {} (y3)
    (u4) edge node {} (y3)
    (u4) edge node {} (y4)
    (u3) edge node {} (v2)
    (u3) edge node {} (u4)
 
    (u3) edge node {} (v2)
    
    (v2) edge node {} (y1)
    (v2) edge node {} (z1)
    (v2) edge node {} (z2)
    (v2) edge node {} (z4)
    (v2) edge node {} (v3)
    (v2) edge node {} (v1)
    (v2) edge node {} (z3)
    (v2) edge node {} (y3)
    
    (z4) edge node {} (v4);
\end{tikzpicture}
\caption{Is $u_2$ or $u_4$ more similar to $u_3$?}
\label{fig:example}
\end{figure}

The ultimate goal for node embeddings is to represent the structural roles of nodes by fixed size vectors which reflect the similarity between nodes. However, there are different notions of node similarity. As an example, consider the toy graph in Figure~\ref{fig:example}. It seems that $u_4$ is similar to $u_3$ because they are part of the same local clique. But on the the other hand one can argue that $u_1$ is more similar to $u_3$ than $u_4$ because they are all directly connected to a hub node like $v_2$. The goal of the present work is to design scalable embedding algorithms that can handle different similarity objectives and have rigorously understood properties. 
%
%
%
%

The main contributions of the paper are as follows:
\begin{itemize}[leftmargin=*]
\item {\bf Theoretical insights.} We formally define the problem of coordinated local node sampling and present novel discrete embedding algorithms that, unlike previous works based on heuristics, {\em provably} preserve the similarity between nodes with respect to different objectives. Furthermore, the algorithms are highly efficient.
 
\item {\bf Scalable model training.} We show how to use the discrete embeddings in scalable kernel models. More precisely, we design methods that approximate the Hamming kernel by explicit feature maps such that we can train a linear support-vector machine for node classification. Previous works require the computation of a Gram matrix which is unfeasible for massive graphs.
\end{itemize}

\paragraph{Organization of the paper} 
\label{sec:org}
In the next section we present notation and outline the problem setting. In Section~\ref{sec:cologne} we present three algorithms for local neighborhood sampling according to different objectives. We analyze the computational complexity of each approach and the properties of the generated samples. In Section~\ref{sec:expl_map} we present an approach to the generation of explicit feature maps for the Hamming kernel, thus enabling the use of discrete node embeddings in scalable kernel models.  We discuss related work in Section~\ref{sec:previous}. An experimental evaluation is presented in Section~\ref{sec:exp}. The paper is concluded in Section~\ref{sec:concl}.

\section{Notation and overview of techniques} \label{sec:overview}

The input is a graph $G = (V, E)$ over $n=|V|$ nodes and $m=|E|$ edges. The distance $d(u,v)$ between nodes $u$ and $v$ is the minimum number of edges that need to be traversed in order to reach $u$ from $v$, i.e., the shortest path from $u$ to $v$. We consider undirected graphs, thus $d(u, v) = d(v, u)$. Also, we assume connected graphs, thus $d(u, v) < \infty$ for all $u, v \in V$. These assumptions are however only for the ease of presentation, all algorithms work for directed or disconnected graphs. The $k$-hop neighbors of node $u$ is the set of nodes $N_k(u) = \{v \in V: d(u, v) \le k\}$. The set of neighbors of node $u$ is denoted as $N(u)$. We call the subgraph induced by $N_k(u)$ the local $k$-hop neighborhood of $u$. 
A {\em discrete embedding vector} is a fixed size vector whose entries come from a discrete set. For example, by sampling with replacements $\ell$ nodes from the $k$-hop neighborhood of each node we create embeddings that consists of other nodes. Or attributes of the neighboring nodes, if existent.   
 
We say $\tilde{q}$ is an $1\pm \varepsilon$-approximation of a quantity $q$ if $(1-\varepsilon)q \le \tilde{q} \le (1+\varepsilon)q$.
%
%

{\bf Sketch based coordinated sampling.}
Our algorithms build upon sketching techniques for massive data summarization. In a nutshell, sketching replaces a vector ${\bf x} \in \mathbb{R}^n$ by a compact data structure $sketch_{\bf{x}} \in \mathbb{R}^d$, $d \ll n$, that approximately preserves many properties of the original $\bf{x}$.
In coordinated sampling~\cite{coord_sampling}, given a universe of elements $U$, and a set of sets $\{S_i \subseteq U\}$, the goal is to draw samples from $U$ such that each set $S_i$ is represented by a compact summary $sketch_{S_i}$. It should hold $\texttt{sim}(sketch_{S_i}, sketch_{S_j}) \approx \texttt{sim}(S_i, S_j)$, i.e., the summaries approximately preserve the similarity between the original sets, for different similarity measures. 

{\bf $L_p$ sampling.}
The $p$-norm of vector $x \in \mathbb{R}^n$ is $\|x\|_p = (\sum_{i=1}^n |x_i|^p)^{1/p}$ for $p \in \mathbb{N}\cup \{0\}$. \footnote{The $0$-norm,  the number of nonzero coordinates, is not a norm in the strict mathematical sense but the notation has become standard.} 
We call $L_p$ sampling a sampling procedure that returns each coordinate $x_i$ from $x$ with probability $\frac{|x_i|^p}{\|x\|_p^p}$.


{\bf Work objective.} 
Let $\mathbf{f}^k_u$ be the $k$-hop frequency vectors of node $u$ such that $\mathbf{f}^k_u[z]$ is the number of unique paths of length at most $k$ from $u$ to $z$. 
Let $s_u \in N_k(u)$ be the node returned by an algorithm $\mathcal{A}$ as a sample for node $u$. We say that $\mathcal{A}$ is a {\em coordinated sampling algorithm} with respect to a similarity measure $\texttt{sim}: V \times V \rightarrow [0,1]$ iff $$\Pr[s_u = s_v] = \texttt{sim}(\mathbf{f}^k_u, \mathbf{f}^k_v) \text{ for } u, v \in V$$


The goal of our work is the design of scalable algorithms for coordinated $L_p$ sampling from local node neighborhoods with rigorously understood properties. We can also phrase the problem in graph algebra terms. Let $A \in \{0,1\}^{n\times n}$ be the adjacency matrix of the graph. The objective is to implement coordinated $L_p$ sampling from each row of $ M_k = \sum_{i=0}^kA^i$  {\em without} explicitly generating the $A^i$.  
By sketching the local $k$-hop neighborhood frequency vector $\mathbf{f}^k_u$ of each node $u$ we will design efficient $L_p$ sampling algorithms. 

\begin{algorithm}
\caption{The overall structure of the coordinated local neighborhood sampling approach.}  \label{alg:main_alg} 
\begin{algorithmic}[0]
\REQUIRE Graph $G = (V, E)$
\FOR{each $u \in V$}
\STATE Initialize $sketch_u$ with node $u$ 
\ENDFOR
\FOR{$i=1$ to $k$}
\FOR{each $u \in V$}
\STATE Update $sketch_u$ by merging all $sketch_v$ for $v \in N(u)$ into $sketch_u$ 
\ENDFOR
\ENDFOR
\FOR{$u \in V$}
\STATE  Return a node from $sketch_u$ as a sample for $u$.
\ENDFOR
\end{algorithmic}
\end{algorithm}

\section{\algname} \label{sec:cologne}

The general form of our approach is in Figure~\ref{alg:main_alg}. We first initialize a sketch at each node $u$ with the node $u$ itself. For $k$ iterations, for each node we collect the sketches from its neighbors and aggregate them into a single sketch. At the end we sample from each node's sketch. The algorithm is used to generate a single coordinate of the embedding vector of each node. For each coordinate we will use a different random seed.

Consider a trivial example. We initialize $sketch_u$ with a sparse $n$-dimensional binary vector such that $sketch_u[u]=1$ is the only nonzero coordinate for all $u \in V$. Merging the sketches is simply entrywise vector addition. We can formally show that after $k$ iterations $sketch_u$ is exactly the $k$-hop frequency vector of $u$, i.e., $sketch^{(k)}_u[v] = \mathbf{f}^{k}_u[v]$. We need to address the following issues: i)~Storing the entire frequency vectors $\mathbf{f}_u^k$ is very inefficient. Even for smaller values of $k$, we are likely to end up with dense $n$-dimensional sketches as most real graphs have a small diameter. ii)~How can we get coordinated samples from the different sketches?

\subsection{Coordinated uniform ($L_0$) sampling}

We first present a simple coordinated sampling algorithm for generating samples from the local neighborhood of each node. The approach builds upon {\em min-wise independent permutations}~\cite{minwise}, a technique for estimating the Jaccard similarity between sets. Assume we are given two sets $A \subseteq U$ and $B \subseteq U$, where $U$ is a universe of elements, for example all integers. We want to estimate the fraction $\frac{|A\cap B|}{|A\cup B|}$. Let $\pi: U \rightarrow U$ be a random permutation of the elements in $U$. With probability $\frac{|A\cap B|}{|A\cup B|}$ the smallest element in $A \cup B$ with respect to the total order defined by $\pi$ is contained in $A\cap B$. The indicator variable denoting whether the smallest elements in $\pi(A)$ and $\pi(B)$ are identical is an unbiased estimator of the Jaccard similarity  $J(A, B)$. The mean of $t = O(\frac{1}{\alpha \varepsilon^2})$ such estimators is an $1\pm \varepsilon$-approximation of $J(A, B) \ge \alpha$ with high probability. (If $J(A, B) < \alpha$ then the additive error is bounded by $\alpha \varepsilon$.) 

An algorithm for sampling uniformly from the $k$-hop neighborhood of node $u$ easily follows, see Algorithm~\ref{alg:l1_alg}. We implement a random permutation on the $n$ nodes by generating a random number for each node $r: V \rightarrow \{0,1,\ldots,\ell-1\}$. For a sufficiently large $\ell$ with high probability $r$ is a bijective function and thus it implements a random permutation\footnote{For $\ell = n^2/\delta$ with probability $1-\delta$ the function is bijective.}. For each node $u$, $sketch_u$ is  initialized with $(r(u), u)$, i.e., the sketch is just a single {\em (random number, node)} pair. The aggregation after each iteration is storing the pair with the smallest random number from $u$'s neighbors, $sketch_u = \min_{(r_v, v): v \in N(u)}sketch_v$. After $k$ iterations at each node $u$ we have the smallest number from the set $\{r(v): v \in N_k(u)\}$, i.e., we have sampled a node from $N_k(u)$ according to the permutation defined by the function $r$. The samples for any two nodes $u$ and $w$ are coordinated as we work with the same permutation on the set $N_k(u) \cup N_k(w) \subseteq V$. The next theorem is a straightforward corollary from the main result on minwise-independent permutations~\cite{minwise}:

\begin{thmx}
For all nodes $u \in V$, we can sample $s_u \in N_k(u)$ with probability $1/|N_k(u)|$  in time $O(mk)$ and space $O(m)$. For any pair of nodes $u, v$ it holds $$\Pr[s_u=s_v] = \frac{|N_k(u) \cap N_k(v)|}{|N_k(u) \cup N_k(v)|}$$
\end{thmx} \label{thm:l0}
\begin{proof}
We show by induction that after $k$ iterations at each node $u$ we have the node $v \in N_k(u)$ with the smallest random value $r(v)$. The statement is clearly true for $k=0$. Assume that for each $v\in N(u)$ we have the smallest value $r(w)$ for $w \in N_{k-1}(v)$. Let $\text{min}_k(u)$ be the smallest node random value for the nodes in $N_k(u)$. Observe that $N_k(u) = \bigcup_{v \in N(u)} N_{k-1}(v)$ and thus $\text{min}\{r(w): w \in N_k(u)\} = \text{min}\{\text{min}_{k-1}(v): v \in N(u)\}$. The theorem then easily follows from the main result from~\cite{minwise}.
\end{proof}

In terms of linear algebra, the above algorithm is an efficient implementation of the following na\"ive approach: Let $A$ be the the adjacency matrix of $G$. Randomly permute the columns of $M_k = \sum_{i=0}^kA^i$, and for each row in the updated $M_k$ select the first nonzero coordinate. Thus, the algorithm implements coordinated $L_0$ sampling from each row of $M_k$ but avoids the explicit generation of $M_k$. 


\subsection{Coordinated $L_p$ sampling} \label{sec:l0}

\begin{algorithm}
\caption{$L_0$ and $L_1$ sampling for single coordinate.}  \label{alg:l1_alg} 
\medskip
{\sc $L_0$ Sampling}
\algsetup{indent=1em}
\begin{algorithmic}[0]
\REQUIRE Graph $G = (V, E)$, random function $r: V \rightarrow (0, 1]$
\FOR{each $u \in V$}
\STATE Initialize $sketch_u$ with $(r(u), u)$ 
\ENDFOR
\FOR{$i=1$ to $k$}
\FOR{each $u \in V$}
\FOR{each $v \in N(v)$}
\STATE  $sketch_u = \min(sketch_u, sketch_v)$ 
\ENDFOR
\ENDFOR
\ENDFOR
\FOR{$u \in V$}
\STATE  Return the node $z$ from the pair $sketch_u = (r(z), z)$
\ENDFOR
\end{algorithmic}
%
\medskip
{\sc $L_1$ Sampling}
\algsetup{indent=1em}
\begin{algorithmic}[0]
\REQUIRE Graph $G = (V, E)$, random function $r: V \rightarrow (0, 1]$, int $\ell$
\FOR{each $u \in V$}
\STATE Initialize $sketch_u$ with node, weight pair $(u, w(u) = 1/r(u))$ 
\ENDFOR
\FOR{$i=1$ to $k$}
\FOR{each $u \in V$}
\FOR{each $v \in N(v)$}
\STATE  Update $sketch_u$ with $sketch_v$, keeping the top $\ell$ (node, weight) pairs.
\ENDFOR 
\ENDFOR
\ENDFOR
\FOR{$u \in V$}
\STATE  Return  node $z$ from $sketch_u$ if $w(z) \ge \|\bf{f}^k_u\|_1$ 
\ENDFOR
\end{algorithmic}
\end{algorithm}

The solution for uniform sampling is simple and elegant but it does not fully consider the graph structure. It only considers if there is a path between two nodes $u$ and $v$ but, unlike in random walks, not how many paths there are between $u$ and $v$. We design coordinated sampling algorithms such that easily accessible nodes are more likely to be sampled. 

Let us present an approach to $L_p$ sampling from data streams for $p \in (0, 2]$~\cite{lp_sampling}. Let $\mathcal{S}$ be a data stream of pairs $i, w_i$ where $i$ is the item and $w_i \in \mathbb{R}$ is the weight update for item $i$, for $i \in [n]$. For example, the network traffic for a website where user $i$ spends $w_i$ seconds at the site. The objective is $L_p$ sampling from the frequency vector of the stream, i.e., return each item $i$ with probability roughly $|\mathbf{f}[i]|^p/\|\mathbf{f}\|^p_p$, where $\mathbf{f}[i] = \sum_{(i, w_i) \in \mathcal{S}}w_i$. (Items can occur multiple times in the stream.) The problem has a simple solution if we can afford to store the entire frequency vector $\mathbf{f}$.  
The solution in~\cite{lp_sampling} is to reweight each item by scaling it by a random number $1/r_i^{1/p}$ for a uniform random $r_i \in (0, 1]$. Let $z_i = \mathbf{f}[i]/r_i^{1/p}$ be the new weight of item $i$. The crucial observation is that $$\Pr[z_i \ge \|\mathbf{f}\|_p] = \Pr[r_i \le {\mathbf{f}[i]^p}/{\|\mathbf{f}\|_p^p}] = \frac{\mathbf{f}[i]^p}{\|\mathbf{f}\|_p^p}$$  One can show that with constant  probability there exists a unique item $i$ with $z_i \ge \|\mathbf{f}\|_p$. Thus, if we know the value of $\|\mathbf{f}\|_p$, a space-efficient solution is to keep a sketch data structure from which we can detect the heavy hitter that will be the sampled item. Estimating the $p$-norm of a frequency vector for $p\in (0,2]$ is a fundamental problem with known solutions~\cite{count_sketch}.

There are two challenges we need to address when applying the above approach to local graph neighborhood sampling.
First, how do we get {\em coordinated} samples? 
Second, if we explicitly generate all entries in the frequency vector of the local neighborhood this would result in time complexity of $O(\sum_{i=0}^knnz(A^i))$,  $nnz(A)$ denoting the number of non-zero elements in the adjacency matrix $A$.  

For the first issue, we achieve coordinated sampling by reweighting the nodes $u \in V$ by $1/r_u$ for random numbers $r_u \in (0, 1]$ for $u \in V$. 
An inefficient solution works then as follows. For each node $u$ we generate a random number $r_u \in (0, 1]$, initialize $\mathbf{w}^0_u[u] = 1/r_u$ and set $\mathbf{w}^0_u[v] = 0$ for all $v \neq u$.  We iterate over the neighbor nodes and update the vector $\mathbf{w}^k_u = \mathbf{w}^{k-1}_u \oplus \sum_{v \in N(u)} \mathbf{w}^{k-1}_v$ where $\oplus$ denotes entrywise vector addition.  We sample a node $v$ iff $\mathbf{w}_u[v] \ge \|\mathbf{f}^{k}_u\|_1$. (Observe that $\|\mathbf{f}^{k}_u\|_1$ can be computed exactly by the same iterative procedure.) 

Our $L_1$ sampling solution in Algorithm~\ref{alg:l1_alg} is based on the idea of {\em mergeable sketches}~\cite{mergeable} such for any vectors $x, y \in \mathbb{R}^n$ it holds $sketch(x+y) = sketch(x) + sketch(y)$. Following the \algname template from Algorithm~\ref{alg:main_alg}, we  iteratively update the sketches at each node. The sketch collected in the $i$-th iteration at node $u$ results from merging the sketches in $N(u)$ at iteration $i-1$, and summarizes the $i$-hop neighborhood $N_i(u)$.  \cite{lp_sampling} use the CountSketch data structure~\cite{count_sketch}. But a CountSketch is just an array of counters from which the frequency of an item can be estimated. This wouldn't allow us to efficiently retrieve the heavy hitter from the sketch as we would need to query the sketch for all $k$-hop  neighborsof each node. (In the setting in~\cite{lp_sampling} a single vector is being updated in a streaming fashion. After preprocessing the stream we can afford to query the sketch for each vector index as this wouldn't increase the asymptotic complexity of the algorithm.)

Here comes the main algorithmic novelty of our approach. Observing that in our setting all weight updates are strictly positive, we design a solution by using another kind of summarization algorithms for frequent items mining, the so called {\em counter based algorithms}~\cite{frequent}. In this family of algorithms, the sketch consists of an explicitly maintained list of frequent items candidates. For a sufficiently large yet compact sketch a heavy hitter is guaranteed to be in the sketch, thus the sampled node will be the heaviest node in the sketch. The algorithm is very efficient as we only need to merge the compact sketches at each node in each iteration.

The above described algorithm can be phrased again in terms of the adjacency matrix $A$. Let $M_k=\sum_{i=0}^kA^i\cdot R^{-1}$ where $R \in \mathbb{R}^{n\times n}$ is a diagonal matrix with diagonal entries randomly selected  from $(0, 1]$. Then from the $i$-th row we return as sample the index $j$ with the largest value $M_{i, j}$ for which the sampling condition $M_{i, j} \ge \|A^{(k)}_{i,:}\|_1$ is satisfied. 
Sampling in this way is coordinated because the $j$-th column of $A^{(k)}$ is multiplied by the same random value $1/r_j$, thus a node $j$ with a small $r_j$ is more likely to be sampled. 

The formal proof for the next theorem is rather technical but at a high level we show that with high probability we can efficiently detect a node that satisfies the sampling condition using a sketch with $O(\log n)$ entries at each node and can be provably detected by an efficient exact heavy hitter mining algorithm. 
\begin{thmx} Let $G$ be a graph over $n$ nodes and $m$ edges, and let $\mathbf{f}^k_u$ be the frequency vector of the $k$-hop neighborhood of node $u \in V$.
For all $u \in V$, we can sample a node $s_u \in N_k(u)$ with probability $\frac{\mathbf{f}^k_u[s_u]}{\|\mathbf{f}^k_u\|_1}$  in time $O(mk\log n)$  and space $O(m + n \log n)$. For each pair of nodes $u, v \in V$ $$\Pr[s_u=s_v] = \sum_{x \in V}\min(\frac{\mathbf{f}^k_u[x]}{\|\mathbf{f}^k_u\|_1}, \frac{\mathbf{f}^k_v[x]}{\|\mathbf{f}^k_v\|_1})$$
\end{thmx} \label{thm:l1}
\begin{proof}
We will show that with constant probability there exists exactly one node $x \in N_k(u)$ that satisfies the sampling condition for node $u$, i.e., $\mathbf{w}^k_u[x] \ge t$ for $t =  \|\mathbf{f}^k_{u}\|_1$ where $\mathbf{w}_u^k$ is the reweighted frequency vector $\mathbf{f}_u^k$. Denote this event by $\mathcal{E}_1$.
Consider a fixed node $x \in N_k(u)$.
First, node $x$ is reweighted by $r_x \sim U(0,1)$, therefore the probability that node $x$ satisfies the sampling condition is $$\Pr[\mathbf{f}^k_{u}[x]/r_x \ge t] = \frac{\mathbf{f}^k_{u}[x]}{\|\mathbf{f}^k_{u}\|_1}$$ 
For fixed $x \in N_k(u)$, let $\mathcal{E}^u_x$ be the event that $x$ is the only node that satisfies the sampling condition and for all other nodes $v \neq x$ it holds $\mathbf{w}^k_u[x] \le t/2$. We lower bound the probability for $\mathcal{E}^u_x$ as follows:
\begin{align}
\Pr[\mathcal{E}^u_x] =  \frac{\mathbf{f}^k_{u}[x]}{\|\mathbf{f}^k_{u}\|_1} \prod_{v \in N_k(u), v \neq x}(1-  \frac{2\mathbf{f}^k_{u}[v]}{t}) \ge \nonumber \\
\frac{\mathbf{f}^k_{u}[x]}{\|\mathbf{f}^k_{u}\|_1} \prod_{v \in N_k(u), v \neq x}\alpha(1-2/t)^{\mathbf{f}^k_{u}[v]} \ge  \texttt{(for $\alpha \in (0, 1]$)} \nonumber \\
\frac{\mathbf{f}^k_{u}[x]}{\|\mathbf{f}^k_{u}\|_1} \alpha (1-  2/t)^{\|\mathbf{f}^k_{u}\|_1} =  
\frac{\mathbf{f}^k_{u}[x]}{\|\mathbf{f}^k_{u}\|_1} e^{ -2\|\mathbf{f}^k_{u}\|_1/t} = \nonumber \\
e^{-2}\frac{\mathbf{f}^k_{u}[x]}{\|\mathbf{f}^k_{u}\|_1} \nonumber
\end{align}

The first inequality follows by observing that $(1-k/n)^n \rightarrow e^{-k}$ and $(1-1/n)^{kn}\rightarrow e^{-k}$ for large $n$, and since both $1-k/n$ and $1-1/n$ are in (0,1), for any fixed $n$ and $k < n$ there must exist a constant $\alpha \in (0,1)$ such that $1-k/n \ge \alpha(1-1/n)^k$.
The second inequality holds because $a^{\sum_{v \in V: v \neq x} w_v} > a^{\sum_{v \in V} w_v}$ for $a \in (0, 1)$ and $w_v > 0$.
\\
For $\mathcal{E}_1$ we observe that the events $\mathcal{E}^u_x$, $x \in N_k(u)$ are pairwise disjoint. $$\Pr[\mathcal{E}_1] = \sum_{x \in N_k(u)} \Pr[\mathcal{E}^u_x] \ge \sum_{x \in  N_k(u)} \frac{\mathbf{f}^k_{u}[x]}{e^2\|\mathbf{f}^k_{u}\|_1} = \Omega(1)$$
\\
Next we show  how to efficiently detect the unique element $x$ for which it holds $\mathbf{f}^k_{u}[x]/r_x \ge \|\mathbf{f}^k_{u}\|_1$. 
With probability at least $1-1/n$ it holds $r_x \in [1/n^2, 1]$ for all $x \in V$. We obtain for the expected value of $\mathbf{w}^k_u[x]=\mathbf{f}^k_{u}[x]/r_x$: 
\begin{align}
\mathbb{E}[\mathbf{f}^k_{u}[x]/r_x] = 
O(\mathbf{f}^k_{u}[x] \int_{1}^{n^2} \frac{1}{t} dt) = 
 O(\mathbf{f}^k_{u}[x]\log n) \nonumber
\end{align}
By linearity of expectation $\mathbb{E}[\|\mathbf{w}_u^k\|_1]={O}(\|\mathbf{f}^k_{u}\|\log n)$. Since the random numbers $r_x$ are independent by Hoeffdings's inequality we obtain that $\|\mathbf{w}_u^k\|_1 = {O}(\|\mathbf{f}^k_{u}\|\log n)$ almost surely. Note that $\|\mathbf{w}_u^k\|_1$ can be computed exactly at each node. Thus, we have shown that there exists a unique node $x$ with weight $\|\mathbf{f}^k_{u}\|_1$ and the total weight of the nodes in $N_k(u)$ is bounded by ${O}(\|\mathbf{f}^k_{u}\|_1\log n)$. Using a deterministic frequent items mining algorithm like {\sc Frequent}~\cite{frequent} we can detect this unique heavy hitter using space $O(\log n)$. Since for all other nodes $v \neq x$ it holds $\mathbf{w}^k_u[x] \le \|\mathbf{f}^k_{u}\|_1/2$, by the main result from~\cite{frequent} it follows that for a summary size of $> 2 \log n$ the heavy hitter will be the only node whose weight in the summary will be at least $\|\mathbf{f}^k_{u}\|_1/2$.  Note that the summaries generated by {\sc Frequent} are mergeable~\cite{mergeable}, and can be merged in time proportional to the number of elements stored in the summary. In each iteration we need to perform exactly $m$ such summary merges, thus each iteration over all nodes takes $O(m\log n)$ and needs space $O(n \log n)$  

To complete the proof consider the probability that $x \in  N_k(u) \cap N_k(v)$ is sampled for both nodes $u$ and $v$, i.e., $s_u=s_v=x$. As shown above, with probability $\Omega(1)$ we have generated samples for $u$ and $v$.   Observe that $x=s_u$ if $r_x \le \mathbf{f}^k_{u}[x]/\|\mathbf{f}^k_{u}\|_1$, thus we have  
\begin{align}
\Pr[x = s_u \wedge x=s_v] = \Pr[\mathcal{E}_x^u \wedge \mathcal{E}_x^v]  = \nonumber \\
  \Pr[r_x \le \mathbf{f}^k_{u}[x]/\|\mathbf{f}^k_{u}\|_1 \wedge r_x \le 
\mathbf{f}^k_{v}[x]/\|\mathbf{f}^k_{v}\|_1] = \nonumber \\
\min(\frac{\mathbf{f}^k_u[x]}{\|\mathbf{f}^k_u\|_1}, \frac{\mathbf{f}^k_v[x]}{\|\mathbf{f}^k_v\|_1}) \nonumber
\end{align}
The $\mathcal{E}^u_x$ events are disjoint for $x \in N_k(u)$. Also, nodes $x \notin  N_k(u) \cap N_k(v)$ contribute 0 to the similarity as they are not reachable by at least one of $u$ or $v$. Summing over $x \in  N_k(u) \cap N_k(v)$ completes the proof. 
\end{proof}

For $L_2$ sampling we can apply the same approach as for $L_1$ sampling. The only difference is that we need to reweight each node $u$ by $1/r_u^{1/2}$ for $r_u \in (0,1]$ and compare the weight of the sample candidate with $\|\mathbf{f}_u^k\|_2$.
However, unlike $\|\mathbf{f}_u^k\|_1$, we cannot compute exactly $\|\mathbf{f}_u^k\|_2$ during the $k$ iterations.  But we can efficiently {\em approximate} the 2-norm of a vector revealed in a streaming fashion, this is a fundamental computational problem for which algorithms with optimal complexity have been designed~\cite{count_sketch}. We show the following result: 
 
\begin{thmx} Let $G$ be a graph over $n$ nodes and $m$ edges, and let $\mathbf{f}^k_u$ be the frequency vector of the $k$-hop neighborhood of node $u \in V$.
For all $u \in V$, we can sample a node $s_u \in N_k(u)$ with probability $(1\pm \varepsilon)\frac{\mathbf{f}^k_u[s_u]^2}{\|\mathbf{f}^k_u\|_2^2}$  in time $O(mk(1/\varepsilon^2 + \log n))$  and space $O(m + n (1/\varepsilon^2 + \log n))$, for a user-defined $\varepsilon \in (0, 1)$. For each pair of nodes $u, v \in V$ $$\Pr[s_u=s_v] = (1\pm \varepsilon)\sum_{x \in V}\min(\frac{\mathbf{f}^k_u[x]^2}{\|\mathbf{f}^k_u\|_2^2}, \frac{\mathbf{f}^k_v[x]^2}{\|\mathbf{f}^k_v\|_2^2})$$
\end{thmx}  \label{thm:l2}
\begin{proof}
The same proof as for $L_1$ sampling holds except for computing exactly the norm $\|\mathbf{w}_u^k\|_2$. Instead we will estimate it using {\sc CountSketch}~\cite{count_sketch}. This is a linear data structure that maintains an array with counters. The critical property of linear sketches is that for each node it holds $$sketch(\sum_{v \in N_k(u)} \mathbf{f}_v) = \sum_{v \in N_k(u)} sketch(\mathbf{f}_v)$$ Using a sketch with $O(1/\varepsilon^2)$ counters we obtain a $1\pm \varepsilon$ approximation of $\|\mathbf{w}_u^k\|_2$ for a user-defined $\varepsilon \in (0, 1)$. Thus, we can recover the heavy hitters in the local neighborhood of each node using the approach from the proof of Theorem 3. Observing that for any $\varepsilon \in (0, 1)$ there exists a constant $c$ such that $\frac{1}{1\pm \varepsilon} \in [1-c\cdot \varepsilon, 1+ c\cdot \varepsilon]$, we obtain the stated  bounds on the sampling probability. 
\end{proof}

The similarity measures preserved in the above two theorems may appear non-intuitive. Below we discuss that in a sense they approximate versions of the well-known {\em cosine similarity}.

\subsection*{Discussion} 

Let us provide some intuition for the similarity measures approximated by the samples in Theorem~2 and Theorem~3. In $L_0$ sampling we treat all nodes in the local neighborhood equally, while in $L_1$ sampling the sampling probability is proportional to the probability that we reach a local neighbor by a random walk. $L_2$~sampling is biased towards high-degree local neighbors, i.e., if a node is reachable by many paths in the local neighborhood then it is even more likely to be sampled. 

The similarity function approximated by $L_0$ sampling is the Jaccard similarity between node sets in the local neighborhood. But the similarity for $L_1$ and $L_2$ sampling is less intuitive. Consider two vectors $x, y \in \mathbb{R}^n$. It holds $$0 \le \sum_{i=1}^n \min(\frac{x[i]^2}{\|x\|_2^2}, \frac{y[i]^2}{\|y|_2^2}) \le \sum_{i=1}^n \frac{x[i]^2}{2\|x\|_2^2} + \sum_{i=1}^n  \frac{y[i]^2}{2\|y|_2^2} = 1$$

In particular, if two nodes share no nodes in their $k$-hop neighborhoods the similarity is 0 and if they have identical frequency vectors the similarity is 1.

Also, for $a\ge 0, b\ge 0$ it holds $\min(a^2, b^2) \le ab$, thus we have that $$\sum_{i=1}^n \min(\frac{x[i]^2}{\|x\|_2^2}, \frac{y[i]^2}{\|y|_2^2}) \le \sum_{i=1}^n \frac{x[i]y[i]}{\|x\|_2 \|y\|_2} = \cos(x, y)$$

On the other hand, assume that $\|x\|_2 \le \|y\|_2$ and for all $i$ it holds $x[i] \le y[i] \le c x[i]$ for some $c>1$ and $\frac{x[i]}{\|x\|_2} \le \frac{y[i]}{\|y\|_2}$.  Then 
$$\sum_{i=1}^n  \min(\frac{x[i]^2}{\|x\|_2^2}, \frac{y[i]^2}{\|y|_2^2}) \ge \sum_{i=1}^n \frac{x[i]y[i]/c}{\|x\|_2 \|y\|_2} = \cos(x, y)/c$$

Thus, the measure in a sense approximates cosine similarity. And for $L_1$ sampling we obtain that the related measure is the so called sqrt-cosine similarity~\cite{sqrt_cos} $$\text{sqrt-cos}(x, y) = \sum_{i=1}^n \frac{x[i]y[i]}{\sqrt{\|x\|_1} \sqrt{\|y\|_1}} $$

\subsection*{Semi-streaming model of computation}
The semi-streaming paradigm was introduced as a computational model for the computation on massive graphs~\cite{semistream}. In this model, an algorithm is allowed to use memory of the order $O(n \cdot \text{polylog}(n))$ but $o(m)$, i.e. sublinear in the number of edges. The assumption is that the graph is stored on an external device, or edges are generated on the fly, and the graph edges can be accessed in a sequential manner. Observe that \algname can be implemented in $k$ passes over the edges where we will keep in memory only the node sketches. Further, the edges can be provided in each pass in arbitrary order, thus the approach works in the semi-streaming model of computation.

\section{An explicit map for the Hamming kernel} \label{sec:expl_map}

A typical application for node embeddings is node classification. Previous works~\cite{nethash,nodesketch} have proposed to train a kernel machine with a Hamming kernel. The Hamming kernel is defined as the overlap of vectors $x, y \in \mathbb{U}^d$ for a universe $\mathbb{U}$: $H(x, y) = \sum_{i=1}^d \mathbbm{1}(x_i=y_i)$, i.e. the number of positions at which $x$ and $y$ are identical. However, this approach requires the explicit generation of a precomputed Gram matrix with $t^2$ entries, where $t$ is the number of training examples. 

A solution would be to represent the embeddings by sparse explicit vectors and train a highly efficient linear SVM~\cite{linear_svm} model. For a universe size $N$, we can represent each vector by a binary $Nd$-dimensional vector $b$ with exactly $d$ nonzeros. But even a linear SVM would be unfeasible as we will likely end up with dense decision vectors with $O(Nd)$ entries. We can use TensorSketch~\cite{tensorsketch}, originally designed for the generation of explicit feature maps for the polynomial kernel. Mapping the discrete vectors to lower dimensional sketches with $O(1/\varepsilon^2)$ entries would preserve the Hamming kernel with an additive error of $\varepsilon d$. Observing that $d$ can be a rather small constant, we present a simple algorithm that is more space-efficient than TensorSketch for $d<1/\varepsilon$. The algorithm hashes each nonzero coordinate of the explicit $Nd$-dimensional vector $b$ to a vector $f(b)$ with $d/\varepsilon$ coordinates. If there is a collision, i.e., $f(b)[i] = f(b)[j]$ for $i \neq j$, we set $f(b)[i]=1$ without adding up $f(b)[i]$ and $f(b)[j]$ as in TensorSketch. We show the following result: 
\begin{thmx}
Let $\mathbb{U}$ be a discrete space and $x, y \in \mathbb{U}^d$ for $d \in \mathbb{N}$. For any $\varepsilon \in (0, 1]$ there is a mapping $f: \mathbb{U}^d \rightarrow \{0,1\}^D$ such that $D=\lceil d/\varepsilon \rceil$ and $H(x, y) = f(x)^Tf(u) \pm \varepsilon d$ with probability 2/3. 
\end{thmx} \label{thm:map}
\begin{proof}
Let $N=|\mathbb{U}|$ be the dimensionality of the universe from which the items in the explicit maps are drawn, e.g., the number of graph nodes.
Let further $m_x, m_y \in \{0,1\}^{dn}$ be the na\"ive explicit feature maps of $x$ and $y$. We hash each nonzero index in $[Nd]$ to a random index in $[d\varepsilon]$. 
First observe that by construction each collision of indices, i.e., $f(i) = f(j)$ for $i\neq j$, can lead to an additive error of exactly one in the inner product $m_x^Tm_y$. Indeed, we either miss a nonzero coordinate on which $m_x$ and $m_y$ agree, or we artificially create such a coordinate because of the collision. Since we don't add up the counters, we won't introduce an error larger than 1 due to a collision.

Thus, we have to bound the number of such collisions. Fix a coordinate $i$. Using that our hash function is uniform and that $\Pr[f(j) = k | f(i)=k] = \Pr[f(j)=k]$, the probability that another fixed index $j$ maps to the same location, i.e. $f(i)=f(j)$, is $\varepsilon/d$. Given that there are $d^2$ nonzero coordinate pairs in total, the expected number of such collisions is bounded by $\varepsilon d$. Implementing the hash function using tabulation hashing~\cite{tab_hashing_orig} and using a result from~\cite{tab_hashing}, the number of collisions is concentrated around the expected value with high probability.
\end{proof}

\section{Related work} \label{sec:previous}



{\bf Random walk based embeddings.}
Pioneering approaches to learning word embeddings, such as word2vec~\cite{word2vec} and GloVe~\cite{glove},  have served as the basis for graph representation learning. 
%
The main idea is that for each graph node $u$, we learn how to predict $u$'s occurrence from its context~\cite{deepwalk}. In natural language the context of each word is the set of surrounding words in a sequence, and for graph nodes the context is the set of local neighbors, thus random walks have been used to generate node sequences. Various algorithms have been proposed that allow some flexibility in selecting local neighbors according to different criteria  \cite{line,node2vec,app,verse}.

{\bf Matrix factorization.}
A branch of node embeddings algorithms work by factorization of (powers of) the adjacency matrix of the graph~\cite{hope,arope}. These algorithms have well-understood properties but can be inefficient as even if the adjacency matrix is usually sparse, its powers can be dense, e.g., the average distance between any two users in the Facebook graph is only 4~\cite{4degrees}. The computational complexity is improved using advanced techniques from linear algebra. 

{\bf Deep learning.}
Node embeddings can be also learned using graph neural networks~\cite{graphsage}. 
GNNs are {\em inductive} and can be applied to previously unseen nodes, while the above discussed approaches, including ours, are {\em transductive} and work only for a fixed graph. This comes at the price of the typical deep learning disadvantages such as slow training and the need for careful hyperparameter tuning. 

{\bf Coordinated local sampling.} \label{sec:nodesketch}
Approaches close to ours are NetHash~\cite{nethash} and NodeSketch~\cite{nodesketch}. NetHash uses minwise hashing to produce embeddings for  attributed graphs. However, it builds individual rooted trees of depth $k$ for each node and its complexity is $O(nt(m/n)^k)$ where $t$ is the maximum number of attributes per node. \algname needs time $O(nt)$ to get the attribute with the minimum value for each node and thus the total complexity is $O(nt+ mk)$. For $k>1$ this is asymptotically faster by a factor of $t(m/n)^{k-1}$.

\begin{figure*}

\center{
\definecolor{myblue}{RGB}{80,80,160}
\definecolor{mygreen}{RGB}{80,160,80}
\definecolor{myorange}{RGB}{255,127,0}


\begin{tikzpicture}[thick,
	scale=0.7,
	transform shape,
  fsnode/.style={fill=myblue},
  ssnode/.style={fill=green},
  osnode/.style={fill=myorange},
  every fit/.style={ellipse,draw,inner sep=-2pt,text width=1.6cm},
  -,shorten >= 1pt,shorten <= 1pt
]
\node [fill=red,yshift=0cm,xshift=3cm,label=below:$u$](u)[] {};

\node [fill=yellow,yshift=1cm,xshift=-5.5cm,label=below:$v_1$](v1)[] {};
\node [fill=yellow,yshift=1cm,xshift=-4cm](v2)[] {};
\node [fill=yellow,yshift=1cm,xshift=-2.5cm](v3)[] {};
\node [fill=yellow,yshift=1cm,xshift=-1cm](v4)[] {};
\node [fill=yellow,yshift=1cm,xshift=0.5cm](v5)[] {};
\node [fill=yellow,yshift=1cm,xshift=2cm](v6)[] {};
\node [fill=yellow,yshift=1cm,xshift=3.5cm](v7)[] {};
\node [fill=yellow,yshift=1cm,xshift=5cm](v8)[] {};
\node [fill=yellow,yshift=1cm,xshift=6.5cm](v9)[] {};
\node [fill=yellow,yshift=1cm,xshift=8cm](v10)[] {};
\node [fill=yellow,yshift=1cm,xshift=9.5cm](v11)[] {};
\node [fill=yellow,yshift=1cm,xshift=11cm,label=below:$v_{12}$](v12)[] {};

\node [fill=red,yshift=2cm,xshift=3cm,label=below:$z$](z)[] {};

\draw (u) -- (v1);
\draw (u) -- (v2);
\draw (u) -- (v3);
\draw (u) -- (v4);
\draw (u) -- (v5);
\draw (u) -- (v6);
\draw (u) -- (v7);
\draw (u) -- (v8);
\draw (u) -- (v9);
\draw (u) -- (v10);
\draw (u) -- (v11);
\draw (u) -- (v12);

\draw (z) -- (v1);
\draw (z) -- (v2);
\draw (z) -- (v3);
\draw (z) -- (v4);
\draw (z) -- (v5);
\draw (z) -- (v6);
\draw (z) -- (v7);
\draw (z) -- (v8);
\draw (z) -- (v9);
\draw (z) -- (v10);
\draw (z) -- (v11);
\draw (z) -- (v12);

\node [fill=blue!50!white,yshift=3cm,xshift=-6cm,label=below:$w_1$](w1)[] {};
\node [fill=blue!50!white,yshift=3cm,xshift=-5cm](w2)[] {};
\node [fill=blue!50!white,yshift=3cm,xshift=-4cm](w3)[] {};
\node [fill=blue!50!white,yshift=3cm,xshift=-3cm](w4)[] {};
\node [fill=blue!50!white,yshift=3cm,xshift=-2cm](w5)[] {};
\node [fill=blue!50!white,yshift=3cm,xshift=-1cm](w6)[] {};
\node [fill=blue!50!white,yshift=3cm,xshift=0cm](wx)[] {};
\node [fill=blue!50!white,yshift=3cm,xshift=1cm](w7)[] {};
\node [fill=blue!50!white,yshift=3cm,xshift=2cm](w8)[] {};
\node [fill=blue!50!white,yshift=3cm,xshift=3cm](w9)[] {};
\node [fill=blue!50!white,yshift=3cm,xshift=4cm](w10)[] {};
\node [fill=blue!50!white,yshift=3cm,xshift=5cm](w11)[] {};
\node [fill=blue!50!white,yshift=3cm,xshift=6cm](w12)[] {};
\node [fill=blue!50!white,yshift=3cm,xshift=7cm](w13)[] {};
\node [fill=blue!50!white,yshift=3cm,xshift=8cm](w14)[] {};
\node [fill=blue!50!white,yshift=3cm,xshift=9cm](w15)[] {};
\node [fill=blue!50!white,yshift=3cm,xshift=10cm](w16)[] {};
\node [fill=blue!50!white,yshift=3cm,xshift=11cm](w17)[] {};
\node [fill=blue!50!white,yshift=3cm,xshift=12cm,label=below:$w_{19}$](w18)[] {};

\draw (z) -- (w3);
\draw (z) -- (w4);
\draw (z) -- (w5);
\draw (z) -- (w8);
\draw (z) -- (w10);

\draw (v1) -- (w3);
\draw (v2) -- (w4);
\draw (v5) -- (w5);
\draw (v7) -- (w15);
\draw (v7) -- (w1);
\draw (v5) -- (w2);
\draw (v9) -- (w7);
\draw (v3) -- (w2);
\draw (v2) -- (w3);
\draw (v4) -- (w5);
\draw (v6) -- (w6);
\draw (v5) -- (w2);
\draw (v5) -- (w8);
\draw (v6) -- (w9);
\draw (v6) -- (w1);
\draw (v9) -- (w13);
\draw (v7) -- (w12);
\draw (v7) -- (w7);
\draw (v3) -- (w7);
\draw (v4) -- (w8);
\draw (v11) -- (w10);
\draw (v7) -- (w11);
\draw (v8) -- (w11);
\draw (v12) -- (w11);
\draw (v12) -- (w9);
\draw (v10) -- (w12);
\draw (v11) -- (w9);
\draw (v7) -- (w9);
\draw (v11) -- (w12);
\draw (v3) -- (w1);
\draw (v10) -- (w12);
\draw (v4) -- (w5);
\draw (v5) -- (w4);
\draw (v7) -- (w18);
\draw (v10) -- (w18);
\draw (v8) -- (w17);
\draw (v11) -- (w17);
\draw (v12) -- (w16);
\draw (v7) -- (w16);
\draw (v11) -- (w15);
\draw (v7) -- (w14);
\draw (v10) -- (w14);
\draw (v10) -- (w13);
\draw (v3) -- (w6);
\draw (v3) -- (wx);
\draw (v1) -- (wx);
\draw (v5) -- (wx);

\end{tikzpicture}
}
\caption{NodeSketch~\cite{nodesketch}  might miss that there are many length-2 paths from $u$ to $z$.}
\label{fig:ns}
\end{figure*}
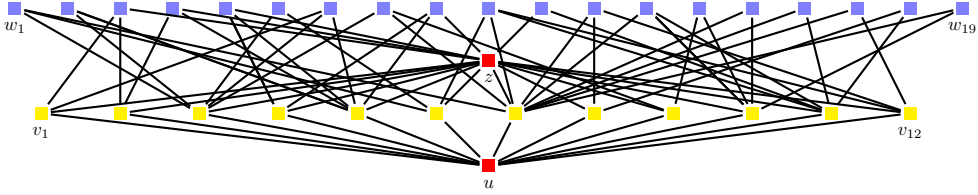

NodeSketch is a highly-efficient {\em heuristic} for coordinated sampling from local neighborhoods. 
It works by recursively sketching the neighborhood at each node until recursion depth $k$ is reached. It builds upon an algorithm for min-max similarity estimation \cite{ioffe}. 
NodeSketch assigns first random weights to nodes, similarly to \algname.  
In the $i$-th recursive call of NodeSketch for each node $u$ we collect samples from $N(u)$.  For node $u$, a {\em single} node from $N(u)$ is selected and this is the crucial difference to \algname. Working with a single node for each iteration is prone to random effects, and the theoretical guarantees from  \cite{ioffe} hold only for sampling from the 1-hop neighborhood.  Consider the graph in Figure~\ref{fig:ns}. There are many paths from node $u$ to node $z$ in $N_2(u)$. In the first iteration of NodeSketch it is likely that most of $u$'s immediate neighbors, the yellow nodes $v_1$ to $v_{12}$, will sample a blue node  as each $v_i$ is connected with many blue $w$ nodes, i.e., $s_{v_i} = w_j$. Thus, in the second iteration it is likely that $u$ ends up with a sample for $u$ different from $z$.  By keeping a sketch \algname   {\em provably} preserves the information that node $z$ is reachable from $u$ by many different paths. And we control the importance we assign to sampling easily reachable nodes by reweighting the nodes by $r_i^{1/p}$.

\section{Experiments} \label{sec:exp}

\begin{table*}[!h]
\footnotesize
\centering
\captionsetup{justification=centering,margin=2cm}
\begin{tabular}{ c  cccccc c cccc}
\multicolumn{1}{c}{{\sc Graph}} &  \multicolumn{6}{c}{{\sc Graph statistics}} &  \multicolumn{5}{c}{{\sc Embedding time} (in seconds)}\\
&&&&&&&&&\\
& nodes & edges & classes & features &  features per node & diameter & $|$ & Random walk & NodeSketch & $L_0$ & $L_1$/$L_2$ \\
\toprule
Cora & 2.7K & 5.4K & 7 & 1.4K & 18.2 & 19 & $|$ & 2.1 & 2.7 & 1.5 & 25.7 \\
\midrule
Citeseer & 3.3K & 4.7K & 6 & 3.7K & 36.1 & 28 & $|$ & 3.3  & 3.8 & 2.3 & 30.4 \\
\midrule
PubMed & 19.7K & 44.3K & 3 & 500 & 50.5 & 18 & $|$ & 17.3  & 35.2 & 15.3 & 214.3  \\
\midrule
Deezer & 28.2K & 92.7K & 2 & 31K & 33.9 & 21 & $|$ & 37.0 & 62.5 & 33.4  & 344.4  \\
\midrule 
LastFM & 7.6K & 27.8K & 18 & 7.8K & 195.3 & 15 & $|$ & 20.4 & 68.6  & 18.9 & 128.5  \\
\midrule
GitWebML & 37.7K & 289K & 2 & 4K & 18.3 & 7 & $|$ & 70.3 & 112.4 & 63.2  & 609.1 \\
\bottomrule
\end{tabular}
\caption{Information on datasets and embedding generation time.} \label{tab:datainfo}
\end{table*}

We evaluate \algname on six publicly available graphs, summarized in Table~\ref{tab:datainfo}. The first three datasets Cora, Citeseer, Pubmed~\cite{sen_et_al} are citation networks where nodes correspond to scientific papers and edges to citations. Each node is assigned a unique class label describing the research topic, and nodes are described by a list of key words. 
LastFM and Deezer represent the graphs of the online social networks of music streaming sites LastFM Asia and Deezer Europe~\cite{lastfm}. The links represent follower relationships and the vertex features are musicians liked by the users. The class labels in LastFM are the users' nationality, and for Deezer -- the users' gender. The GitWebML graph represents a social network where node classes correspond to web or ML developers who have starred at least 10 repositories and edges to mutual follower relationships. Node features are location, starred repositories, employer and e-mail address~\cite{musae}.  

We sample node attributes by using the base sampling algorithm. For example, for each keyword {\em kw} describing the citation network nodes we generate a random number $r_{kw}$ and then at each node we sample keywords according to the corresponding algorithm (e.g., we take the word with smallest $r_{kw}$ for $L_0$ sampling, sample according to min-max sampling for NodeSketch~\cite{nodesketch}, etc.)

We compare \algname against NodeSketch. Our $L_0$ sampling yields identical results to NetHash but, as discussed, our algorithm is much more efficient.  Note that we don't compare with continuous embeddings. Such a comparison can be found in~\cite{nodesketch} but, as argued in the introduction, discrete embeddings offer various advantages such as interpretability and scalability. 

We generated $d=50$ samples from the $k$-hop neighborhood of each node for $k \in \{1,2,3,4\}$, resulting in $50$-dimensional embeddings, using following methods: i) A standard random walk (RW) of length $k$ that returns the last visited node. ii) NodeSketch (NS) as described in~\cite{nodesketch}. iii) \algname for $L_p$ sampling for $p \in \{0,1,2\}$, using a sketch with 10 nodes for $p\in\{1,2\}$.

\paragraph{Embedding generation time for varying embedding size}
In Table~\ref{tab:embtime} we show the embedding generation time for embedding vectors of dimensionality 10, 25, and 50, for the slowest $L_1/L_2$ sampling. We observe almost perfect time increase proportional to the embedding size. 
\\
\vspace*{6mm}
\begin{table}[h!]
\centering
\begin{tabular}{ c ccc}
 & 10 & 25 & 50 \\
 \toprule
 Cora & 9.1 & 17.2 & 33 \\
 Citeseer & 10.1 & 20.6 &  40.8 \\
 Pubmed & 63.9 & 138.9 & 279.6\\
 Deezer & 105 & 225.7 & 442.6 \\
 LastFM & 41.1 & 88.3 & 173.4 \\ 
 GitWebML & 202.4 & 429.1 & 852.3\\
 \bottomrule
\end{tabular}
\vspace{2mm}
\caption{\centering Embedding generation time for $L_1/L_2$ sampling.} \label{tab:embtime}
\end{table}

\begin{table}
\centering
\begin{tabular}{ c cc}
 & kernel SVM & explicit map \\
 \toprule
 Cora & 0.3 & 0.7 \\
 Citeseer & 0.5 & 1.2  \\
 Pubmed & 26.3 & 12.2 \\
 Deezer & 138 & 16.2  \\
 LastFM & 2.1 & 2.4  \\ 
 GitWebML & OOM & 15.9  \\
 \bottomrule
\end{tabular}
\caption{SVM training and inference time for \mbox{NodeSketch} (in seconds).} \label{tab:rtime}
\end{table}

\begin{table*}
\hspace*{-0.6cm}
\scriptsize
\begin{tabular}{ll  c c c c c} 
\multicolumn{2}{c}{} &  \multicolumn{1}{c}{RandomWalk} &  \multicolumn{1}{c}{NodeSketch} &  \multicolumn{1}{c}{$L_0$} &  \multicolumn{1}{c}{$L_1$} &  \multicolumn{1}{c}{$L_2$}\\
\toprule
\multirow{4}{*}{Cora}   & Accuracy & 0.454 $\pm$ 0.013 \ (hop 1) &{\bf 0.833 $\pm$ 0.016 \ (hop 1)} &0.824 $\pm$ 0.018 \ (hop 1) & {\em 0.831 $\pm$ 0.018 \ (hop 1)} &0.829 $\pm$ 0.019 \ (hop 1)\\
 & Balanced accuracy & 0.393 $\pm$ 0.015 \ (hop 1) & {\bf 0.820 $\pm$ 0.020 \ (hop 1)} & 0.808 $\pm$ 0.023 \ (hop 1) & {\em 0.814 $\pm$ 0.023 \ (hop 1)} &0.812 $\pm$ 0.023 \ (hop 1)\\
 & Micro AUC & 0.811 $\pm$ 0.008 \ (hop 1) &0.972 $\pm$ 0.006 \ (hop 1) &0.970 $\pm$ 0.006 \ (hop 1) &{\bf 0.974 $\pm$ 0.005 \ (hop 1)} &{\em 0.973 $\pm$ 0.005 \ (hop 1)}	\\
 & Macro AUC & 0.786 $\pm$ 0.008 \ (hop 1) &{\em 0.969 $\pm$ 0.006 \ (hop 1)} &0.967 $\pm$ 0.006 \ (hop 1) &{\bf 0.971 $\pm$ 0.005 \ (hop 1)} &{\em 0.969 $\pm$ 0.005 \ (hop 1)} \\
\midrule
\multirow{4}{*}{Citeseer}  & Accuracy & 0.346 $\pm$ 0.016 \ (hop 1) &0.706 $\pm$ 0.013 \ (hop 3) &0.702 $\pm$ 0.017 \ (hop 2) &{ \em 0.711 $\pm$ 0.015 \ (hop 2)} & {\bf 0.722 $\pm$ 0.015 \ (hop 2)}\\
 & Balanced accuracy & 0.318 $\pm$ 0.013 \ (hop 1) &0.672 $\pm$ 0.015 \ (hop 2) &0.668 $\pm$ 0.019 \ (hop 2) &{\em 0.676 $\pm$ 0.017 \ (hop 2)} &{\bf 0.691 $\pm$ 0.016 \ (hop 2)}\\
 & Micro AUC & 0.679 $\pm$ 0.010 \ (hop 2) &0.906 $\pm$ 0.008 \ (hop 2) &0.911 $\pm$ 0.009 \ (hop 1) &{\em 0.915 $\pm$ 0.007 \ (hop 2)} & {\bf 0.917 $\pm$ 0.007 \ (hop 2)}\\
 & Macro AUC & 0.648 $\pm$ 0.012 \ (hop 1) &0.891 $\pm$ 0.010 \ (hop 4) &0.892 $\pm$ 0.008 \ (hop 2) &{\em 0.898 $\pm$ 0.007 \ (hop 2)} & {\bf 0.903 $\pm$ 0.007 \ (hop 2)}\\
\midrule
\multirow{4}{*}{PubMed}  
 & Accuracy & 0.589 $\pm$ 0.008 \ (hop 2) &0.798 $\pm$ 0.006 \ (hop 4) &0.781 $\pm$ 0.005 \ (hop 1) &{\bf 0.818 $\pm$ 0.004 \ (hop 2)} &{\em 0.804 $\pm$ 0.007 \ (hop 4)}\\
 & Balanced accuracy & 0.577 $\pm$ 0.007 \ (hop 2) &0.785 $\pm$ 0.007 \ (hop 4) &0.778 $\pm$ 0.004 \ (hop 1) &{\bf 0.809 $\pm$ 0.004 \ (hop 2)} &{\em 0.793 $\pm$ 0.006 \ (hop 2)} \\
 & Micro AUC & 0.789 $\pm$ 0.004 \ (hop 2) &0.927 $\pm$ 0.003 \ (hop 1) &0.919 $\pm$ 0.002 \ (hop 1) &{\bf 0.938 $\pm$ 0.002 \ (hop 2)} & {\em 0.932 $\pm$ 0.002 \ (hop 2)}\\
 & Macro AUC & 0.778 $\pm$ 0.004 \ (hop 2) &0.924 $\pm$ 0.003 \ (hop 1) &0.915 $\pm$ 0.002 \ (hop 1) &{\bf 0.935 $\pm$ 0.002 \ (hop 2)} & {\em 0.928 $\pm$ 0.002 \ (hop 2)}\\
\midrule
\multirow{4}{*}{DeezerEurope}  & Accuracy & 0.540 $\pm$ 0.006 \ (hop 3) &0.550 $\pm$ 0.007 \ (hop 1) &0.554 $\pm$ 0.003 \ (hop 1) &{\em 0.558 $\pm$ 0.008 \ (hop 2)} & {\bf 0.565 $\pm$ 0.006 \ (hop 2)}\\
 & Balanced accuracy & 0.518 $\pm$ 0.008 \ (hop 2) &0.542 $\pm$ 0.006 \ (hop 1) &0.545 $\pm$ 0.004 \ (hop 1) &{\em 0.548 $\pm$ 0.005 \ (hop 1)} & {\bf 0.556 $\pm$ 0.006 \ (hop 2)}\\
 & Micro AUC & 0.571 $\pm$ 0.004 \ (hop 3) &0.588 $\pm$ 0.005 \ (hop 1) &{\em 0.589 $\pm$ 0.004 \ (hop 1)} &{\em 0.589 $\pm$ 0.006 \ (hop 2)} &{\bf 0.594 $\pm$ 0.005 \ (hop 2)}\\
 & Macro AUC & 0.529 $\pm$ 0.004 \ (hop 3) &0.561 $\pm$ 0.005 \ (hop 1) &0.564 $\pm$ 0.004 \ (hop 1) &{\em 0.565 $\pm$ 0.006 \ (hop 2)} & {\bf 0.577 $\pm$ 0.005 \ (hop 2)}\\
\midrule
\multirow{4}{*}{LastFM}   & Accuracy & 0.424 $\pm$ 0.010 \ (hop 3) &0.809 $\pm$ 0.008 \ (hop 4) &0.725 $\pm$ 0.015 \ (hop 1) & {\bf 0.825 $\pm$ 0.008 \ (hop 4)}  & {\em 0.818 $\pm$ 0.009 \ (hop 4)}\\
 & Balanced accuracy & 0.294 $\pm$ 0.022 \ (hop 3) &{\bf 0.720 $\pm$ 0.021 \ (hop 4)} & 0.62 $\pm$ 0.02 \ (hop 2) &{\em 0.707 $\pm$ 0.025 \ (hop 3)} &0.695 $\pm$ 0.028 \ (hop 4)\\
 & Micro AUC & 0.877 $\pm$ 0.007 \ (hop 3) &{\bf 0.971 $\pm$ 0.003 \ (hop 2)} & 0.966 $\pm$ 0.002 \ (hop 1) &0.969 $\pm$ 0.003 \ (hop 1) &{\em 0.970 $\pm$ 0.003 \ (hop 1)}\\
 & Macro AUC & 0.768 $\pm$ 0.007 \ (hop 3) &0.935 $\pm$ 0.003 \ (hop 1) &0.924 $\pm$ 0.002 \ (hop 1) &{\em 0.936 $\pm$ 0.003 \ (hop 1)} &{\bf 0.939 $\pm$ 0.003 \ (hop 1)}\\
\midrule
\multirow{4}{*}{GitWebML}  
 & Accuracy & 0.811 $\pm$ 0.004 \ (hop 1) &0.823 $\pm$ 0.004 \ (hop 1) &0.831 $\pm$ 0.003 \ (hop 1) &{\em 0.835 $\pm$ 0.003 \ (hop 1)} &{\bf 0.837 $\pm$ 0.003 \ (hop 1)}\\
 & Balanced accuracy & 0.685 $\pm$ 0.005 \ (hop 1) &0.751 $\pm$ 0.006 \ (hop 1) &0.759 $\pm$ 0.006 \ (hop 1) &{\bf 0.764 $\pm$ 0.006 \ (hop 1)} & {\em 0.763 $\pm$ 0.006 \ (hop 1)}\\
 & Micro AUC & 0.851 $\pm$ 0.003 \ (hop 1) &0.899 $\pm$ 0.003 \ (hop 1) &0.904 $\pm$ 0.002 \ (hop 1) &{\bf 0.908 $\pm$ 0.002 \ (hop 1)} &{\em 0.905 $\pm$ 0.002 \ (hop 1)}\\
 & Macro AUC & 0.757 $\pm$ 0.003 \ (hop 1) &0.848 $\pm$ 0.003 \ (hop 1) &0.854 $\pm$ 0.002 \ (hop 1) &{\bf 0.862 $\pm$ 0.002 \ (hop 1)} & {\em 0.855 $\pm$ 0.002 \ (hop 1)}\\
\bottomrule
\end{tabular}
\caption{The best result for each approach for the different hop depths. The overall best result is given in {\bf bold} font, and the second best -- in {\em italics}. }
\label{tab:results}
\end{table*}

\begin{table*}[h!]
\scriptsize
\begin{tabular}{ll  c c c c c} 

\multicolumn{2}{c}{} &  \multicolumn{1}{c}{RandomWalk} &  \multicolumn{1}{c}{NodeSketch} &  \multicolumn{1}{c}{$L_0$} &  \multicolumn{1}{c}{$L_1$} &  \multicolumn{1}{c}{$L_2$}\\
\toprule
\multirow{2}{*}{Cora} & Precision & 0.002 $\pm$ 0.01 \ (hop 3)  & 0.014 $\pm$ 0.003 \ (hop 1) & 0.014 $\pm$ 0.004 \ (hop 1) & {\em 0.016 $\pm$ 0.003} \ (hop 1) &{\bf 0.017 $\pm$ 0.003 \ (hop 1)}\\
 & Recall & 0.037 $\pm$ 0.02 \ (hop 3) &0.266 $\pm$ 0.057 \ (hop 1) &0.271 $\pm$ 0.063 \ (hop 1) & {\em 0.317 $\pm$ 0.045 \ (hop 1)} & {\bf 0.334 $\pm$ 0.041 \ (hop 1)}\\
\midrule
\multirow{2}{*}{Citeseer}   & Precision & $< 0.001$ & {\em 0.020 $\pm$ 0.003 \ (hop 2)} & 0.017 $\pm$ 0.004 \ (hop 4) & 0.020 $\pm$ 0.003 \ (hop 2) & {\bf 0.021 $\pm$ 0.003 \ (hop 2)}\\
 & Recall & 0.015 $\pm$ 0.02 \ (hop 1) & {\em 0.453 $\pm$ 0.041 \ (hop 2)} & 0.385 $\pm$ 0.055 \ (hop 2) & 0.452 $\pm$ 0.045 \ (hop 2) & {\bf 0.491 $\pm$ 0.066 \ (hop 2)}\\
\midrule
\multirow{2}{*}{PubMed}  
  & Precision & 0.001 $\pm$ 0.001 \ (hop 1) & 0.001 $\pm$ 0.001 \ (hop 1) & {\em 0.011 $\pm$ 0.003 \ (hop 1)} &{\bf 0.025 $\pm$ 0.007 \ (hop 1)} & 0.007 $\pm$ 0.003 \ (hop 1)\\
 & Recall & 0.002 $\pm$ 0.002 \ (hop 1) &0.002 $\pm$ 0.002 \ (hop 1) & {\em 0.025 $\pm$ 0.006 \ (hop 1)} & {\bf 0.056 $\pm$ 0.015 \ (hop 1)} & 0.016 $\pm$ 0.007 \ (hop 1)\\
\midrule
\multirow{2}{*}{DeezerEurope}  & Precision & $< 0.001$  & $< 0.001$  & $< 0.001$ & $< 0.001$ & $< 0.001$\\
 & Recall & $< 0.001$ & $< 0.001$ & $< 0.001$ & $< 0.001$ & $< 0.001$ \\
\midrule
\multirow{2}{*}{LastFM}   & Precision & $< 0.001$ & 0.024 $\pm$ 0.005 (hop 1) & 0.031 $\pm$ 0.004 \ (hop 1) &{\em 0.038 $\pm$ 0.004 \ (hop 1)} &{\bf 0.047 $\pm$ 0.005 \ (hop 1)}\\
 & Recall & $< 0.001$ & 0.084 $\pm$ 0.016 \ (hop 1) & 0.112 $\pm$ 0.016 \ (hop 1) &{\em 0.136 $\pm$ 0.016 \ (hop 1)} & {\bf 0.167 $\pm$ 0.018 \ (hop 1)}
\\
\midrule
\multirow{2}{*}{GitWebML}  
 & Precision & $< 0.001$ & 0.001 $\pm$ 0.001 (hop 1) & 0.002 $\pm$ 0.001 (hop 1) & {\bf 0.018 $\pm$ 0.005 (hop 1)} & {\em 0.016 $\pm$ 0.004 (hop 1)}\\
 & Recall & $< 0.001$ & 0.002 $\pm$ 0.002 (hop 1) & 0.003 $\pm$ 0.002 (hop 1) & {\bf 0.031 $\pm$ 0.008 (hop 1)} & {\em 0.027 $\pm$ 0.006 (hop 1)}\\
\bottomrule
\end{tabular}
\caption{The best result (precision@1000 and recall@1000) for each approach for the different hop depths for link prediction. The overall best result is given in {\bf bold} font, and the second best -- in {\em italics}. }
\label{tab:link_prediction}
\end{table*}
\subsection{Embedding evaluation}

\paragraph{Running time.}
The algorithms were implemented in Python 3 and run on a Linux machine with 16 GB main memory and a 4.3 GHz Ryzen 7 CPU. 
In the right half of Table~\ref{tab:datainfo} we present results for the running time of embeddings generation for 50-dimensional embeddings. We observe that $L_0$ sampling and NodeSketch are highly efficient, $L_0$ sampling is even faster than random walks as we need to generate only a single random number per node. $L_1/L_2$ sampling is slower which is due to the fact that we update a sketch at each node. 

\paragraph{Parallelization}

Parallelization of the embedding generation approach is easy. Each coordinate of the embedding vector is generated by running a copy of the \algname algorithm. The pseudocode for the general case in Algorithm~\ref{alg:main_alg}, and the $L_0$ and $L_1$ versions from Algorithm~\ref{alg:l1_alg}, are for sampling a single coordinate of the embedding vector. Clearly, we can run copies of the algorithm in parallel, we only have to ensure that each copy has a unique random seed. For example, in order to generate 50-dimensional embeddings we can run 5 copies of the algorithm such that each of them produces 10-dimensional embeddings. The final embedding vector is then a concatenation of the 10-dimensional embeddings generated by different threads. Note that we can also generate embeddings in a distributed setting. 

We implemented parallelized versions of NodeSketch and $L_1$ sampling.  We used Python's \texttt{multiprocessing} library for parallelization. The results can be seen in Table~\ref{tab:parallel_ns} (NodeSketch) and Table~\ref{tab:parallel_l1} (\algname).

\begin{table}[h!]
\centering
\begin{tabular}{ c ccc}
\# threads & 1 & 5 & 10 \\
 \toprule
 Cora & 2.3 & 0.9 & 0.9 \\
 Citeseer & 3.0 & 1.4 &  1.1 \\
 Pubmed & 30.1 & 9.4 & 8.9\\
 Deezer & 52.6 & 15.4 & 13.8 \\
 LastFM & 55.9 & 16.7 & 14.1 \\ 
 GitWebML & 88.9 & 31.6 & 23.8\\
 \bottomrule
\end{tabular}
\caption{\centering Time for NodeSketch sampling using parallelization for 50-dimensional embeddings generation.} \label{tab:parallel_ns}
\end{table}

\begin{table}[h!]
\centering
\begin{tabular}{ c ccc}
\# threads & 1 & 5 & 10 \\
 \toprule
 Cora & 25.7 & 6.7 & 5.4 \\
 Citeseer & 30.4 & 8.1 &  6.3 \\
 Pubmed & 214.3 & 57.4 & 42.2\\
 Deezer & 344.4 & 93.5 & 62.9 \\
 LastFM & 128.5 & 35.2 & 26.5 \\ 
 GitWebML & 609.1 & 166.3 & 120.0\\
 \bottomrule
\end{tabular}
\caption{\centering Running time for $L_1$ sampling with sketch size 10 using parallelization for 50-dimensional embeddings generation.} \label{tab:parallel_l1}
\end{table}

\begin{figure}
\centering
\includegraphics[scale=0.42]{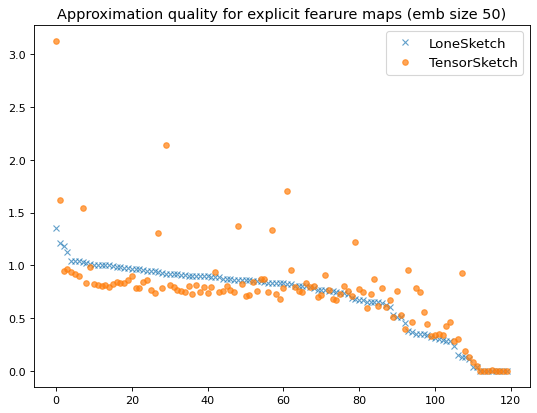}
\caption{Comparison of the approximation quality for our approach from Theorem~4 and TensorSketch.}
\label{fig:appr_cmp}
\end{figure}

\paragraph{Sketch quality.} We set $\varepsilon=0.01$ and as we have $d=50$, the expected error for each inner product is bounded by $0.5$. We observe that the higher variance in TensorSketch leads to outliers. In Figure~\ref{fig:appr_cmp} we show the ratio of the actual error to the expected error for the sketching procedure from Theorem~4, and for TensorSketch~\cite{tensorsketch} for a sample of 1,000 vector pairs for NodeSketch. (The approximation quality is similar for the \algname algorithms.)

\paragraph{Node classification.}
We consider node classification for comparison of the different approaches. We use a linear SVM model with explicit feature maps as presented in Theorem~4, with tabulation hashing which is known to approximate the behavior of truly random hash functions~\cite{tab_hashing}.
We split the data into 80\% for training and 20\% for testing, and use the default $C=1$ as an SVM regularization parameter.

\paragraph{Training and prediction time.} In Table~\ref{tab:rtime} we compare the running time for training and prediction of the linear SVM model as presented in Theorem~4, and an SVM model with precomputed kernel. The embedding vectors are computed by NodeSketch (the values for \algname are similar). Kernel SVMs are considerably slower for larger graphs and for GitWebML result in out-of-memory errors.

\begin{figure}[h!]
\captionsetup{justification=centering,margin=2cm}
\begin{tabular}{ccc}
  \includegraphics[width=50mm]{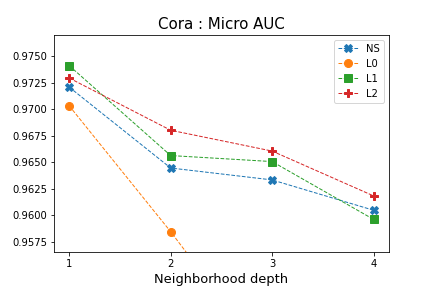} & \includegraphics[width=50mm]{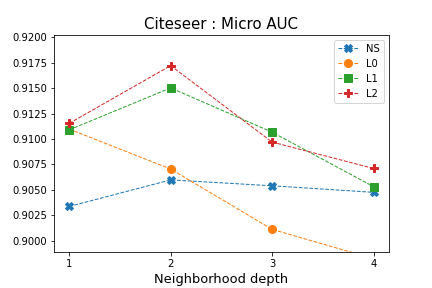} &
   \includegraphics[width=50mm]{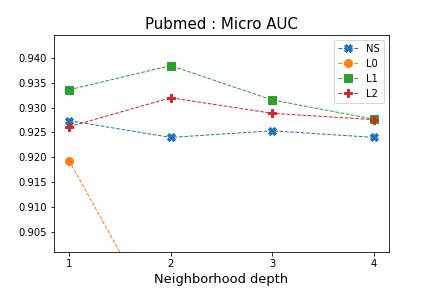} \\ \includegraphics[width=50mm]{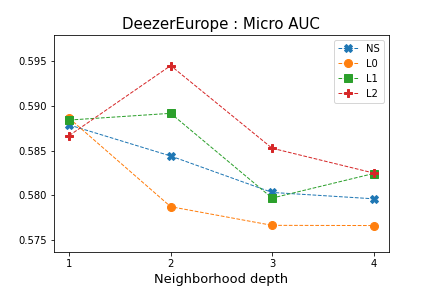} &
    \includegraphics[width=50mm]{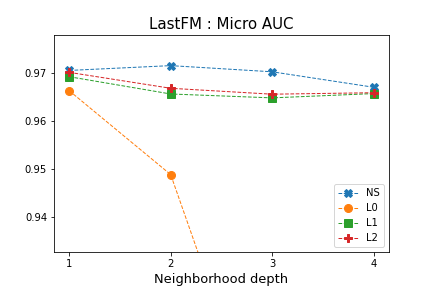} & \includegraphics[width=50mm]{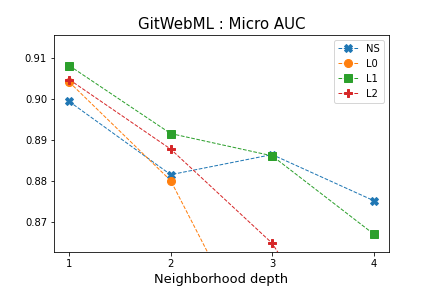} 
\end{tabular}
\caption{Micro AUC scores for the six graphs for varying neighborhood depth.}
\label{fig:clf_micro}
\end{figure}

\begin{figure}[h!]
\captionsetup{justification=centering,margin=2cm}
\begin{tabular}{ccc}
  \includegraphics[width=50mm]{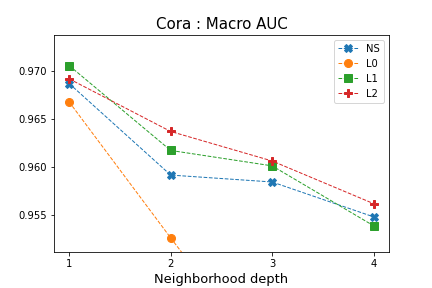} & \includegraphics[width=50mm]{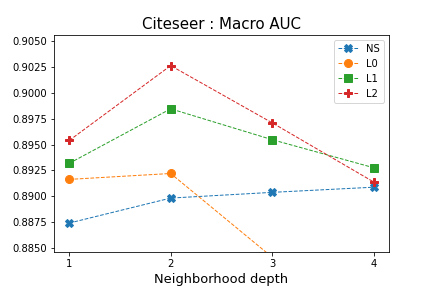} &
   \includegraphics[width=50mm]{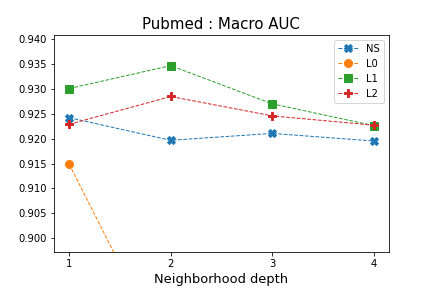} \\ \includegraphics[width=50mm]{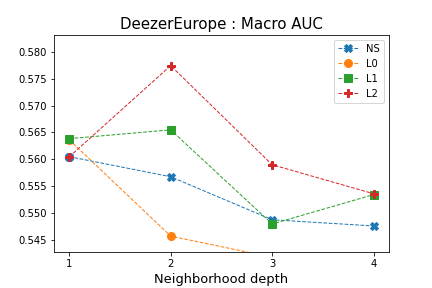} &
    \includegraphics[width=50mm]{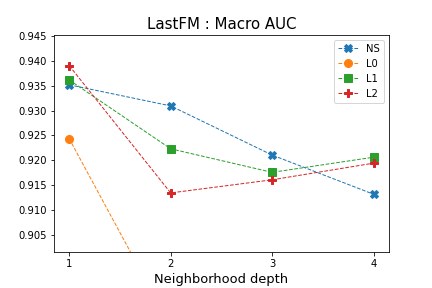} & \includegraphics[width=50mm]{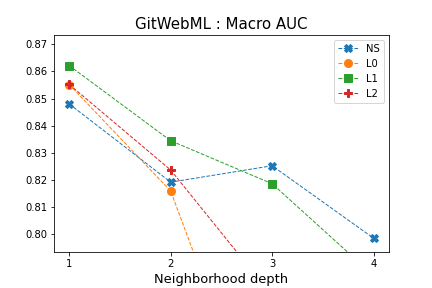} 
\end{tabular}
\caption{Macro AUC score for the six graphs for varying neighborhood depth.}
\label{fig:clf_macro}
\end{figure}

\paragraph{Node classification.}
We evaluate the performance of the algorithms with respect to accuracy, balanced accuracy, micro-AUC, and macro-AUC metrics\footnote{The definitions can be found at  \href{https://scikit-learn.org/stable/modules/classes.html\#module-sklearn.metrics}{https://scikit-learn.org/stable/modules/classes.html\#module-sklearn.metrics}}. We report the mean and standard deviation of 10 independent train/test splits in Table~\ref{tab:results}. We see that overall $L_1$ and $L_2$ sampling achieve the best results. In particular, we observe that $L_0$ sampling yields good results only for a smaller neighborhood depth $k$, this can be clearly seen in the plots in Figures~\ref{fig:clf_micro} and~\ref{fig:clf_macro} where we show the micro-AUC and macro-AUC scores for different neighborhood depths. The reason is that for larger $k$ and a small graph diameter many nodes end up with identical samples: for a connected graph with diameter at most $k$ all nodes will have the same sampled node as their $k$-hop neighborhood is the node set $V$. In contrast, $L_1$ and $L_2$ sampling do not suffer from this drawback as samples depend on the graph structure, not just the set of reachable nodes.  

We provide additional details to the experimental evaluation. In Figures
~\ref{fig:clf_micro}, and~\ref{fig:clf_macro} we plot the micro-AUC and macro-AUC scores for each of the six graphs for the explicit feature map approach. 

%

\paragraph{Link prediction.}
We evaluated the generated sketches on a link prediction task following a similar approach as in~\cite{nodesketch}. We remove 20\% of the edges from each graph such that the graph remains connected. Then we sampled 5\% of all node pairs in the graph. We sorted the sampled pairs in reverse order according to their overlap and considered the 1,000 pairs with the largest overlap. For these 1,000 pairs we computed the precision@1000 and recall@1000. We report the mean and standard deviation for the best neighborhood depth for each approach in Table~\ref{tab:link_prediction}. 
We observe that \algname consistently achieves better results than NodeSketch,  one order of magnitude better for the PubMed and GitWebML graphs. All approaches fail on Deezer because of the large number of node attributes which make the overlap very small.  We observe that the average overlap for the \algname approaches is larger than for NodeSketch, and this has a more pronounced effect for link prediction, see discussion in the next paragraph. The best values are achieved for hop-1 neighborhoods, i.e., sampling from the feature set of the node's immediate neighbors, and this is contrast to the node classification task where the majority of best results for most graphs were achieved when sampling from higher order neighborhoods. (Here, we would like to reiterate that we first create a sketch at each node consisting of the node attributes, and then runs \algname for 1 iteration by merging the neighborhood sketches.) The large overall number of node features in the Deezer graph results again in very low quality predictions. This appears to be a limitation of discrete node embedding algorithms, and one should look for remedies such as reducing the number of features. 

\paragraph{Average overlap}
In Table~\ref{tab:overlap} we show that the \algname approaches have a significantly higher overlap than NodeSketch, for a sample of 1,000 node pairs. This can be also visually observed in Figure~\ref{fig:approx_qual} where we show the approximation quality of the explicit feature maps. Apparently, this has a less significant impact on node classification than on link prediction. In the latter, we create an ordered list of candidate nodes based on the overlap with a query node. Thus, the lower overlap is apparently more likely to lead to random fluctuations and thus more inaccurate link predictions. 

\begin{table}[h!]
\centering
\begin{tabular}{ c cccc}
\#  & NodeSketch & $L_0$ & $L_1$ & $L_2$ \\
 \toprule
 Cora & 1.816 & 8.06 & 8.498 & 7.986 \\
 Citeseer & 1.516 & 3.983 & 4.029 & 3.533 \\
 Pubmed & 1.538 & 27.61 & 15.93 & 15.15\\
 Deezer & 2.286 &6.273 & 8.494 & 8.048 \\
 LastFM & 3.983 & 23.49 & 18.928 & 15.69 \\ 
 GitWebML & 14.65 & 18.02 & 32.364 & 39.509\\
 \bottomrule
\end{tabular}
\caption{\centering Average overlap between node pairs NodeSketch and \algname (hop 2, embedding size 50).} \label{tab:overlap}
\end{table}
\subsection*{Optimal values for the hyperparameters.}
\begin{figure}[h!]
\begin{tabular}{ccc}
  \includegraphics[width=50mm]{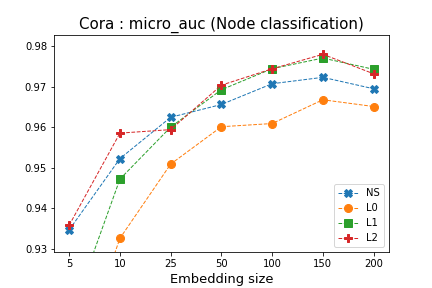} & \includegraphics[width=50mm]{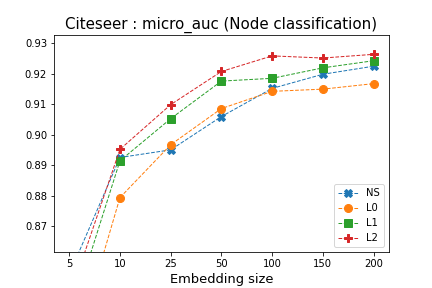} &
   \includegraphics[width=50mm]{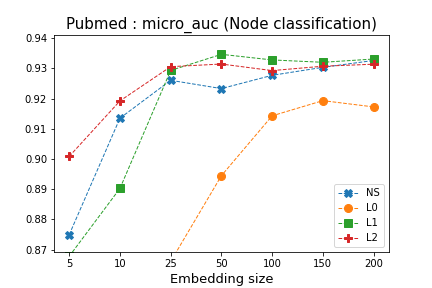} \\ \includegraphics[width=50mm]{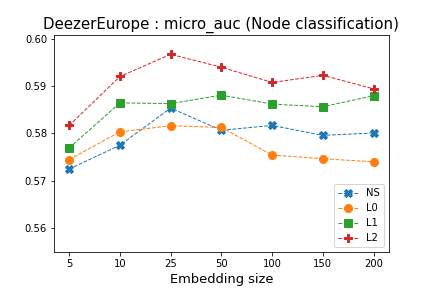} &
    \includegraphics[width=50mm]{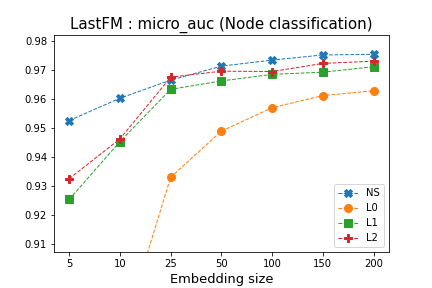} & \includegraphics[width=50mm]{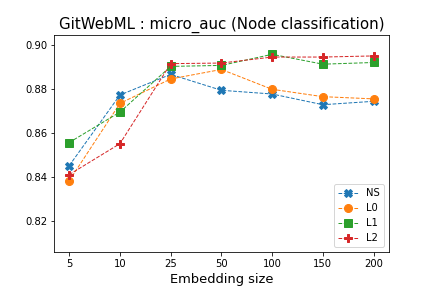} 
\end{tabular}
\caption{Micro AUC scores for varying embedding size.}
\label{fig:clf_micro_var}
\end{figure}

\begin{figure}[h!]
\begin{tabular}{ccc}
  \includegraphics[width=50mm]{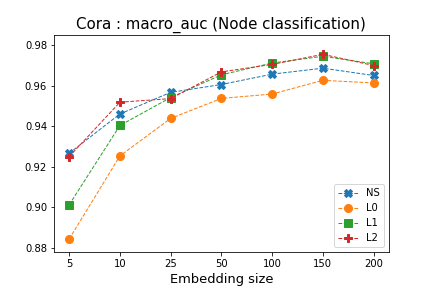} & \includegraphics[width=50mm]{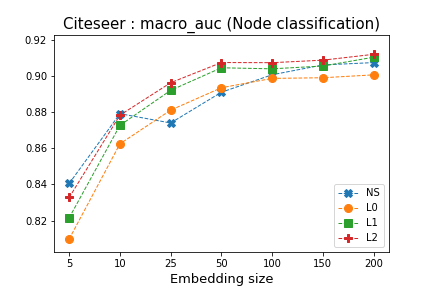} &
   \includegraphics[width=50mm]{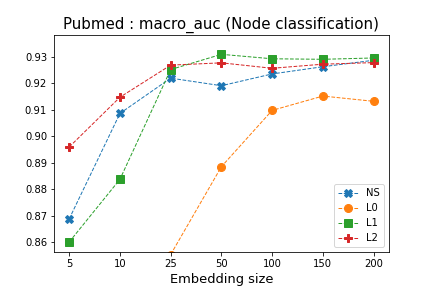} \\ \includegraphics[width=50mm]{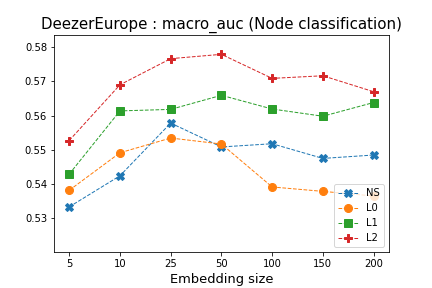} &
    \includegraphics[width=50mm]{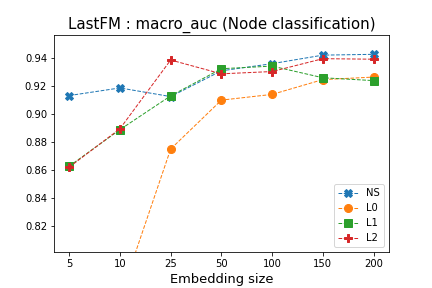} & \includegraphics[width=50mm]{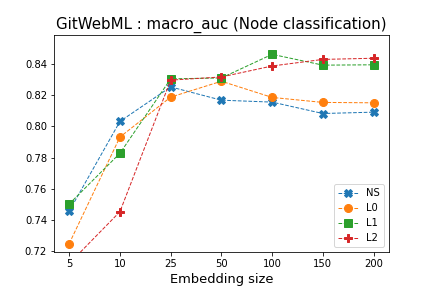} 
\end{tabular}
\caption{Macro AUC scores for varying embedding size for node classification.}
\label{fig:clf_macro_var}
\end{figure}

\begin{figure}[h!]
\begin{tabular}{ccc}
  \includegraphics[width=50mm]{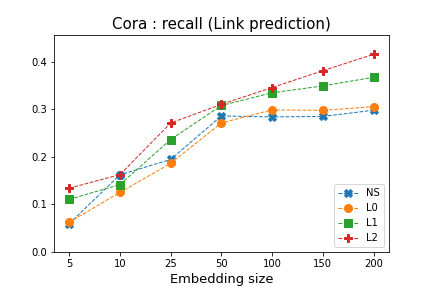} & \includegraphics[width=50mm]{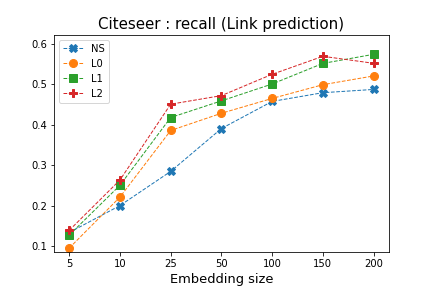} & \includegraphics[width=50mm]{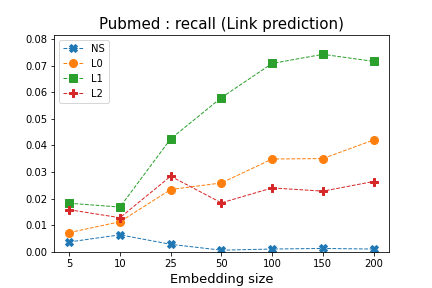} \\  \includegraphics[width=50mm]{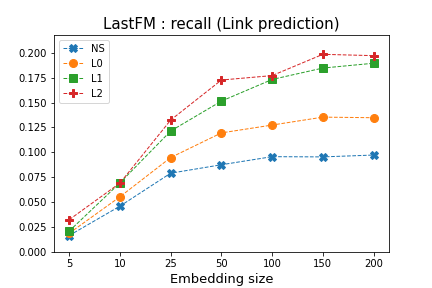} & \includegraphics[width=50mm]{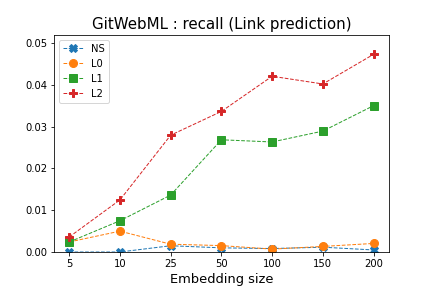} &
\end{tabular}
\caption{Recall@1000 for varying embedding size for link prediction.}
\label{fig:lp_macro_var}
\end{figure}

\begin{figure}[h!]
\begin{tabular}{ccc}
  \includegraphics[width=50mm]{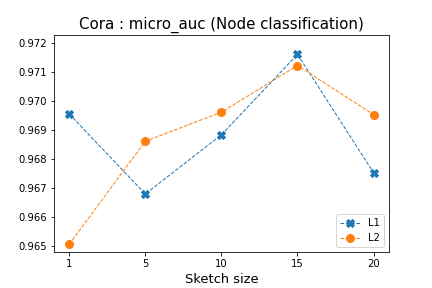} & \includegraphics[width=50mm]{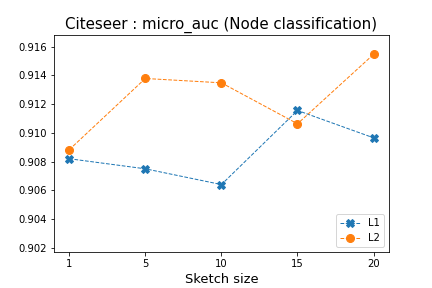} &
   \includegraphics[width=50mm]{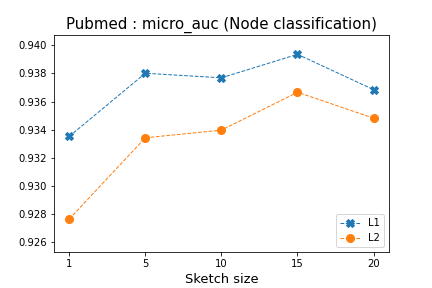} \\ \includegraphics[width=50mm]{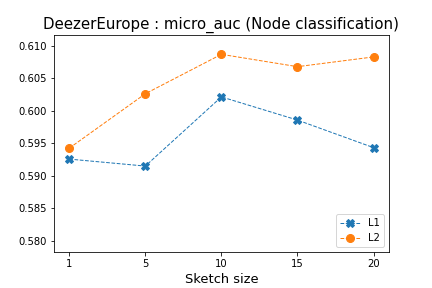} &
    \includegraphics[width=50mm]{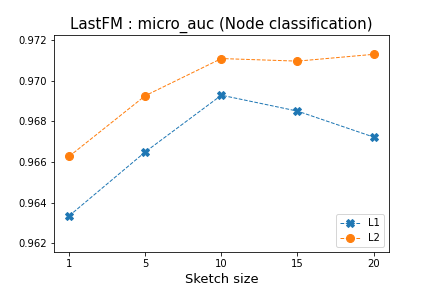} & \includegraphics[width=50mm]{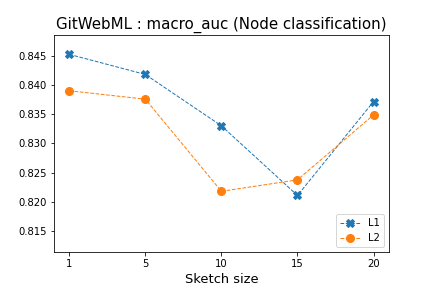} 
\end{tabular}
\caption{Micro AUC scores for varying sketch size for node classification.}
\label{fig:clf_micro_var_ss}
\end{figure}

\begin{figure}[h!]
\begin{tabular}{ccc}
  \includegraphics[width=50mm]{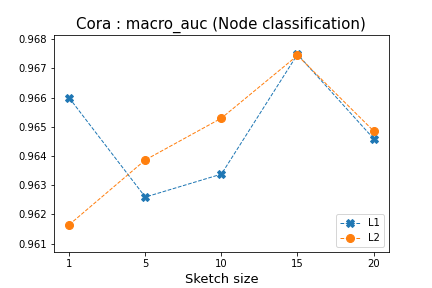} & \includegraphics[width=50mm]{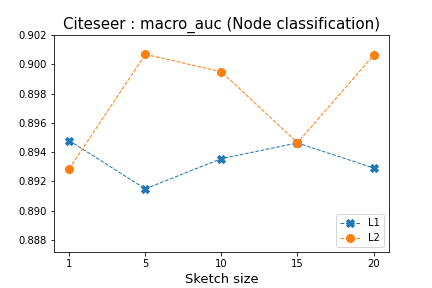} &
   \includegraphics[width=50mm]{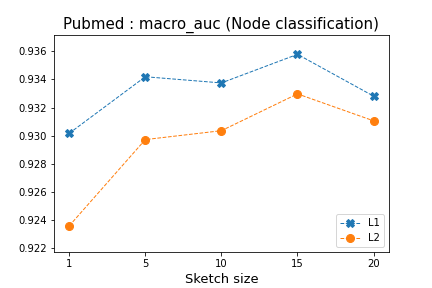} \\ \includegraphics[width=50mm]{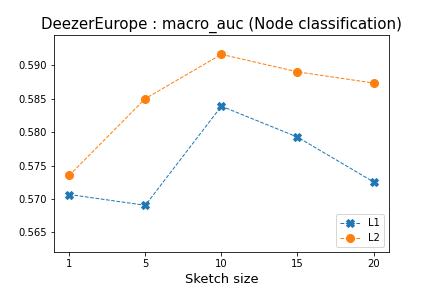} &
    \includegraphics[width=50mm]{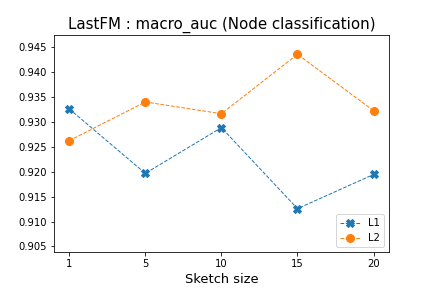} & \includegraphics[width=50mm]{var_sketch_size_GitWebML_50_macro_auc.png} 
\end{tabular}
\caption{Macro AUC scores for varying sketch size for node classification.}
\label{fig:clf_macro_var_ss}
\end{figure}

We analyze the optimal values for the different hyperparameters: the neighborhood depth $k$, the dimensionality of the embeddings $d$, and the sketch size $\ell$ for $L_1/L_2$ sampling. While the optimal $k$ appears to be graph specific, we observe the improvements for larger $d$, see Figures~\ref{fig:clf_micro_var}, \ref{fig:clf_macro_var} and \ref{fig:lp_macro_var}\footnote{The recall scores for the DeezerEurope graph are very low for all methods and we omit the plot from Figure~\ref{fig:lp_macro_var}.}, and the sketch size $\ell$ for $L_1/L_2$ sampling to follow a diminishing returns pattern (Figures~\ref{fig:clf_micro_var_ss}  and \ref{fig:clf_macro_var_ss}). It is worth noting that even for very small embedding dimensions we still achieve very good results for node classification. Also, for larger $d$ NodeSketch catches up with $L_1/L_2$ sampling as more samples compensate for the inconsistencies due to its heuristic nature but this is still not the case for link prediction as the overall overlap remains low and still subject to random fluctuations.

%
%
%
%
%
%
%
%

\begin{table}[h!]
\centering
\begin{tabular}{ c ccc}
\#  & $L_1/L_2$ & DeepWalk & node2vec \\
 \toprule
 Cora & 7.5 & 8.1 & 16.0\\
 Citeseer & 8.5 & 9.2 &  16.7 \\
 Pubmed & 53.4 & 68.3 & 154.2\\
 Deezer & 75.6 & 122.8 & 239.8 \\
 LastFM & 32.3 & 24.2 & 64.1 \\ 
 GitWebML & 145.8 & 206.7 & 1037.8\\
 \bottomrule
\end{tabular}
\caption{\centering Embedding generation time for 50-dimensional embeddings generation (8 cores).} \label{tab:emb_gen_time}
\end{table}

\begin{table*}[h!]
\centering
\footnotesize
\begin{tabular}{ll  c c c c c} 
\multicolumn{2}{c}{} &  \multicolumn{1}{c}{DeepWalk} &  \multicolumn{1}{c}{node2vec} &   \multicolumn{1}{c}{$L_1$} &  \multicolumn{1}{c}{$L_2$}\\
\toprule
\multirow{4}{*}{Cora}   & Accuracy  & 0.802  $\pm$ 0.014 & 0.815 $\pm$ 0.016 & {\em 0.831 $\pm$ 0.018 \ (hop 1)} &0.829 $\pm$ 0.019 \ (hop 1)\\
 & Balanced accuracy & 0.777 $\pm$ 0.013 & 0.804 $\pm$ 0.023 & {\em 0.814 $\pm$ 0.023 \ (hop 1)} &0.812 $\pm$ 0.023 \ (hop 1)\\
 & Micro AUC & 0.954 $\pm$ 0.005 &  0.964 $\pm$ 0.004 &{\bf 0.974 $\pm$ 0.005 \ (hop 1)} &{\em 0.973 $\pm$ 0.005 \ (hop 1)}	\\
 & Macro AUC & 0.946 $\pm$ 0.006 & 0.959 $\pm$ 0.004 &{\bf 0.971 $\pm$ 0.005 \ (hop 1)} &{\em 0.969 $\pm$ 0.005 \ (hop 1)} \\
\midrule
\multirow{4}{*}{Citeseer}  & Accuracy & 0.551 $\pm$ 0.010 & 0.572 $\pm$ 0.018 &{ \em 0.711 $\pm$ 0.015 \ (hop 2)} & {\bf 0.722 $\pm$ 0.015 \ (hop 2)}\\
 & Balanced accuracy & 0.512 $\pm$ 0.011 & 0.522 $\pm$ 0.014 &{\em 0.676 $\pm$ 0.017 \ (hop 2)} &{\bf 0.691 $\pm$ 0.016 \ (hop 2)}\\
 & Micro AUC & 0.828 $\pm$ 0.007 &  0.832 $\pm$ 0.014 &{\em 0.915 $\pm$ 0.007 \ (hop 2)} & {\bf 0.917 $\pm$ 0.007 \ (hop 2)}\\
 & Macro AUC &  0.803 $\pm$ 0.006 &  0.816 $\pm$ 0.0168 &{\em 0.898 $\pm$ 0.007 \ (hop 2)} & {\bf 0.903 $\pm$ 0.007 \ (hop 2)}\\
\midrule
\multirow{4}{*}{PubMed}  
 & Accuracy & 0.792 $\pm$ 0.004 & 0.807 $\pm$ 0.005 &{\bf 0.818 $\pm$ 0.004 \ (hop 2)} &{\em 0.804 $\pm$ 0.007 \ (hop 4)}\\
 & Balanced accuracy & 0.772 $\pm$ 0.005 & 0.790 $\pm$ 0.005 &{\bf 0.809 $\pm$ 0.004 \ (hop 2)} &{\em 0.793 $\pm$ 0.006 \ (hop 2)} \\
 & Micro AUC & 0.917 $\pm$ 0.003 & 0.928 $\pm$ 0.002 &{\bf 0.938 $\pm$ 0.002 \ (hop 2)} & {\em 0.932 $\pm$ 0.002 \ (hop 2)}\\
 & Macro AUC & 0.912 $\pm$ 0.003 & 0.923 $\pm$ 0.002 &{\bf 0.935 $\pm$ 0.002 \ (hop 2)} & {\em 0.928 $\pm$ 0.002 \ (hop 2)}\\
\midrule
\multirow{4}{*}{DeezerEurope}  & Accuracy & 0.556 $\pm$ 0.003 & 0.561 $\pm$ 0.002 &{\em 0.558 $\pm$ 0.008 \ (hop 2)} & {\bf 0.565 $\pm$ 0.006 \ (hop 2)}\\
 & Balanced accuracy & 0.509 $\pm$ 0.004 & 0.522 $\pm$ 0.003 &{\em 0.548 $\pm$ 0.005 \ (hop 1)} & {\bf 0.556 $\pm$ 0.006 \ (hop 2)}\\
 & Micro AUC & 0.581 $\pm$ 0.004 & 0.585 $\pm$ 0.003 &{\em 0.589 $\pm$ 0.006 \ (hop 2)} &{\bf 0.594 $\pm$ 0.005 \ (hop 2)}\\
 & Macro AUC & 0.547 $\pm$ 0.005 & 0.555 $\pm$ 0.003 &{\em 0.565 $\pm$ 0.006 \ (hop 2)} & {\bf 0.577 $\pm$ 0.005 \ (hop 2)}\\
\midrule
\multirow{4}{*}{LastFM}   & Accuracy & {\em 0.857 $\pm$ 0.005} & {\bf 0.866 $\pm$ 0.006} & 0.825 $\pm$ 0.008 \ (hop 4)  & {0.818 $\pm$ 0.009 \ (hop 4)}\\
 & Balanced accuracy & {\em 0.766 $\pm$ 0.007} & {\bf 0.773 $\pm$ 0.007} & 0.707 $\pm$ 0.025 \ (hop 3) &0.695 $\pm$ 0.028 \ (hop 4)\\
 & Micro AUC & {\em 0.974 $\pm$ 0.002} & {\bf 0.975 $\pm$ 0.004} & 0.969 $\pm$ 0.003 \ (hop 1) &{ 0.970 $\pm$ 0.003 \ (hop 1)}\\
 & Macro AUC & {\em 0.943 $\pm$ 0.011} & {\bf 0.944 $\pm$ 0.010} &{ 0.936 $\pm$ 0.003 \ (hop 1)} &{ 0.939 $\pm$ 0.003 \ (hop 1)}\\
\midrule
\multirow{4}{*}{GitWebML}  
 & Accuracy & {\em 0.857 $\pm$ 0.003} &{\bf 0.863 $\pm$ 0.004}  & 0.835 $\pm$ 0.003 \ (hop 1) &{ 0.837 $\pm$ 0.003 \ (hop 1)}\\
 & Balanced accuracy & {\em 0.771 $\pm$ 0.004} & {\bf 0.779 $\pm$ 0.005} &{ 0.764 $\pm$ 0.006 \ (hop 1)} & { 0.763 $\pm$ 0.006 \ (hop 1)}\\
 & Micro AUC & {\em 0.923 $\pm$ 0.002} & {\bf 0.926 $\pm$ 0.003} & 0.908 $\pm$ 0.002 \ (hop 1) & 0.905 $\pm$ 0.002 \ (hop 1)\\
 & Macro AUC & {\em 0.886 $\pm$ 0.003} & {\bf 0.891 $\pm$ 0.003} & 0.862 $\pm$ 0.002 \ (hop 1) & 0.855 $\pm$ 0.002 \ (hop 1)\\
\bottomrule
\end{tabular}
\vspace{1mm}
\caption{The best result for DeepWalk, node2vec and \algname. The overall best result is given in {\bf bold} font, and the second best -- in {\em italics}. }
\label{tab:emb_comparison}
\end{table*}

\subsection*{Comparison with continuous embeddings}

In Table~\ref{tab:emb_gen_time} we show the computational time for each approach. Even the slow $L_1/L_2$ sampling are considerably faster for larger graphs than the two continuous embedding approaches. 

We would like to emphasize that comparing the performance of discrete and continuous embeddings is not really fair. The two embeddings types have very different properties. We want to compare embeddings that look like 

\begin{verbatim}
cell,gene,infection,..,cell,young,dna
\end{verbatim}

with embeddings that look like 
\begin{verbatim}
0.234,-0.67,0.012,..,0.848,-0.703,-0.447
\end{verbatim}

The main advantages of discrete embeddings are their interpetability, easiness to compute, and that do not depend on optimization techniques like gradient descent. On the other hand, continuous embeddings are much more powerful as they can be used by virtually every machine learning algorithm, from logistic regression to deep learning. In the case of discrete embeddings our options are much more limited.  

Having said that, in Table~\ref{tab:emb_comparison} we provide the classification accuracy for the four considered metrics of $L_1$, $L_2$ sampling and DeepWalk~\cite{deepwalk} and node2vec\cite{node2vec} for a Linear SVM with 80\%-20\% train--test split, and report the mean and standard deviation for 10 runs. We use the default parameters provided in the respective libraries, i.e., 10 random walks of length 80 per node, and 5 training epochs. For node2vec we report the best score of all 9 combinations for the hyperparameters $p$ and $q$ from the set \{0.5, 1, 2\}. We observe that \algname achieves better results for the lower density graphs. It is tempting to conjecture that for denser graphs random walks explore more of the neighborhood while the discrete embeddings fail to visit some nodes. 

\begin{figure}[h!]
\centering
\captionsetup{justification=centering}
\begin{tabular}{cc}
  \includegraphics[width=50mm]{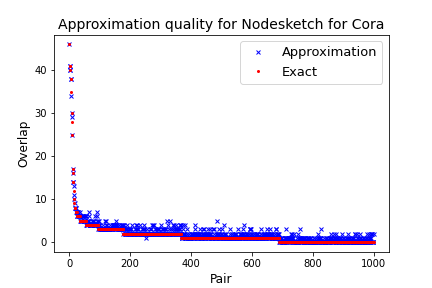} & \includegraphics[width=50mm]{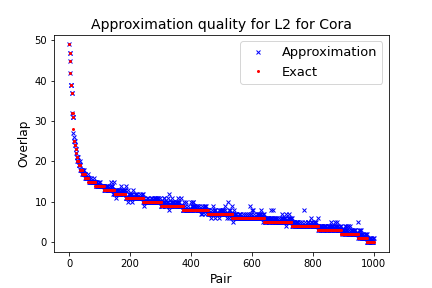} \\
  \includegraphics[width=50mm]{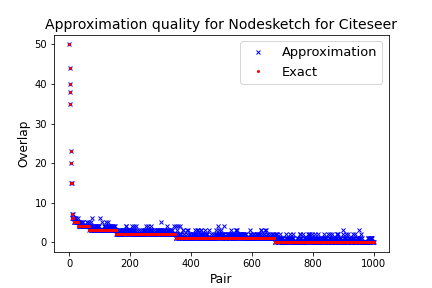} & \includegraphics[width=50mm]{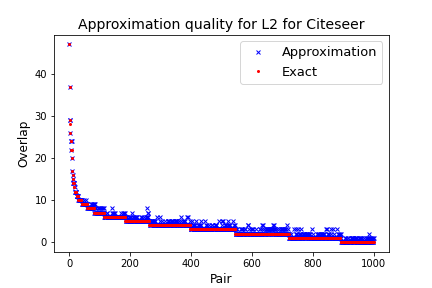} \\
  \includegraphics[width=50mm]{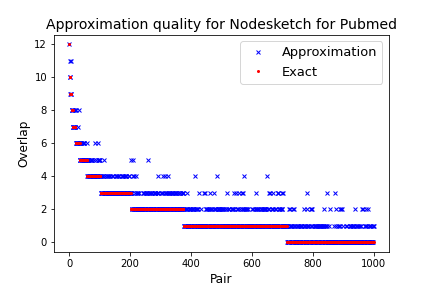} & \includegraphics[width=50mm]{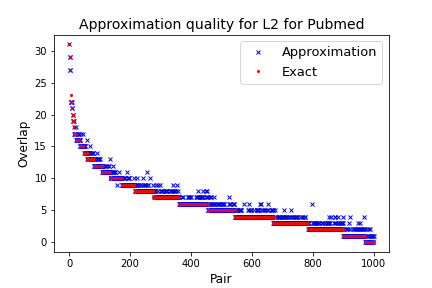} \\
  \includegraphics[width=50mm]{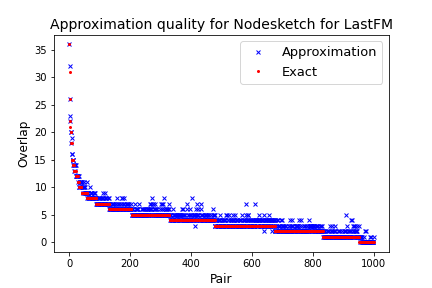} & \includegraphics[width=50mm]{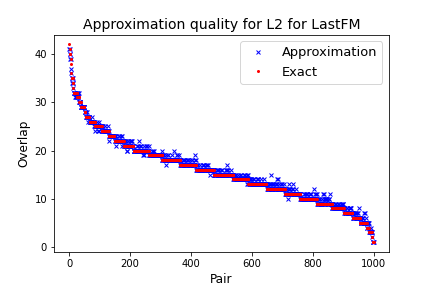} \\
  \includegraphics[width=50mm]{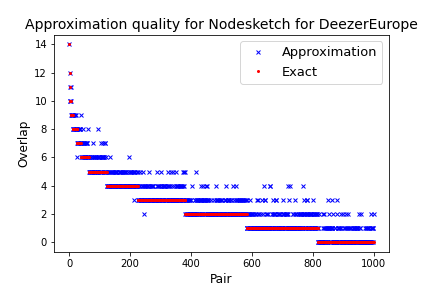} & \includegraphics[width=50mm]{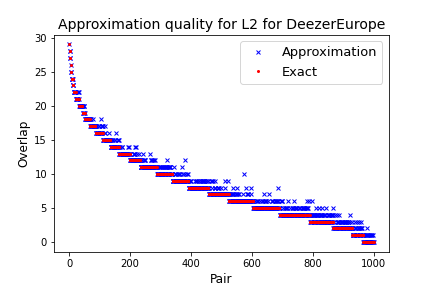} \\
  \includegraphics[width=50mm]{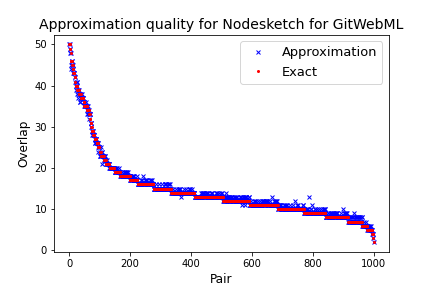} & \includegraphics[width=50mm]{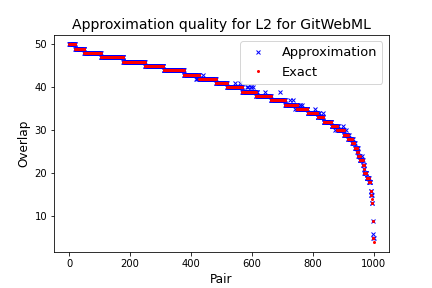} \\
   
\end{tabular}
\caption{ Approximation quality of the overlap kernel.}
\label{fig:approx_qual}
\end{figure}

\paragraph{Approximation guarantee}
In Figure~\ref{fig:approx_qual} we plot the approximation of the Hamming kernel using the explicit map approach from Section~\ref{sec:expl_map} for 10,000 randomly selected node pairs for each of the six graphs, for the NodeSketch and $L_2$ embeddings. One observation is that the overlap for \algname is on average larger than for NodeSketch. This is due to the fact that NodeSketch samples only a single node at each iteration and is susceptible to random fluctuations. Interestingly though, the larger overlap does not always translate into improved classification accuracy but the effects on link prediction are more clearly pronounced.

\subsection*{Embedding interpretability}
Finally, we discuss how discrete node embeddings can be used in interpretable machine learning. Consider the PubMed graph. The nodes represent scientific articles. Each article is described by a list of key words that appear in the paper abstract, these are the node features that we list in Table~\ref{tab:datainfo}. We sample a feature for each node and then propagate the sampled features using \algname or NodeSketch. For example, if the word \texttt{cell} appears in many abstracts in the local neighborhood of a given node, it is likely to be sampled. After sampling $d$ samples, we end up with a vector consisting of key words. An embedding vector can look like 
\begin{verbatim}
[cell, oral, complex, cell, ... , insulin, cell, young, dna]
\end{verbatim} 

We can compare two such embedding vectors by the overlap kernel discussed in Section~\ref{sec:expl_map} and design a kernel SVM model. But we can also use other, more interpretable  approaches. For example, using a model like a decision tree we can explain predictions by looking at the splits of the tree when classifying a new example. (A research paper is classified as belonging to the category ``Genetics'' because when splitting on coordinate 10 it had a value of \texttt{cell}, and then splitting on coordinate 16 the value is \texttt{dna}).

\begin{figure}[h]
\centering
\hspace*{-5mm}
\includegraphics[width=84mm]{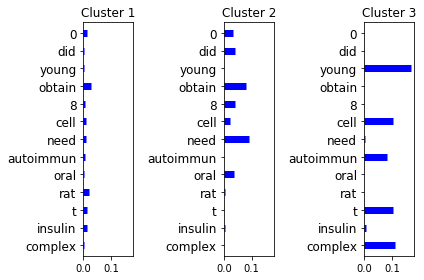}
\caption{Top words distribution in the three different clusters of Pubmed nodes.}
\label{fig:clusters}
\end{figure}

The embeddings can be also used in unsupervised learning. In Figure~\ref{fig:clusters} we show the result of clustering the Pubmed vectors into three clusters, the distance measure is based on the overlap between the discrete embedding vectors from the 2-hop neighborhood using $L_1$ sampling. (This is part of another ongoing project of ours and we include it only in order to address the reviewer's comment.) We plot the distribution of the words (on a log scale) of embeddings in each cluster for the top words and observe the distributions are very different between clusters. We can then divide papers into different groups depending on the vocabulary of their local neighborhood.   

\section{Conclusions and future work} \label{sec:concl}

We presented new algorithms for coordinated local graph sampling with rigorously understood theoretical properties. We demonstrate that using sketching techniques with well-understood properties also has practical advantages and can lead to more accurate algorithms in downstream tasks, and explicit feature maps open the door to highly scalable classification algorithms.  

Finally, we would like to pose an open question. Can we learn structural roles~\cite{struct_roles} by assigning appropriate node attributes? In particular, can we combine the sampling procedure with approaches for graph labeling?

\clearpage
\bibliographystyle{plain}
\bibliography{lone}

\end{document}